\documentclass[10pt,twocolumn,letterpaper]{article}

\usepackage{iccv}
\usepackage{times}
\usepackage{epsfig}
\usepackage{graphicx}
\usepackage{amsmath}
\usepackage{amssymb}
\usepackage{amsmath,bm}
\usepackage{multirow}
\usepackage{booktabs}
\usepackage[table]{xcolor}
\usepackage{amsmath}
\usepackage{amssymb}
\usepackage{color}
\usepackage{times}
\usepackage{epsfig}
\usepackage{booktabs,multirow}
\usepackage{algorithm}
\usepackage{algorithmicx}
\usepackage{algpseudocode}
\usepackage{graphics}
\usepackage{threeparttable}
\usepackage{color}
\usepackage[normalem]{ulem}
\usepackage{multirow}
\usepackage{float}
\usepackage{amsfonts}
\usepackage{bm}
\usepackage{subfig}
\usepackage{enumitem}
\usepackage[numbers]{natbib}
\usepackage{array}
\usepackage[table]{xcolor}
\usepackage{colortbl}
\definecolor{bblue}{rgb}{0,150,230}
\definecolor{mygray}{gray}{.9}
\makeatletter
\newcommand{\thickhline}{%
    \noalign {\ifnum 0=`}\fi \hrule height 1pt
    \futurelet \reserved@a \@xhline
}
\makeatother

\DeclareMathOperator*{\argmax}{argmax}

\usepackage[pagebackref=true,breaklinks=true,colorlinks,bookmarks=false]{hyperref}

\iccvfinalcopy % *** Uncomment this line for the final submission

\def\iccvPaperID{2055} % *** Enter the ICCV Paper ID here

\ificcvfinal\fi
\begin{document}

%%%%%%%%% TITLE
\title{The Ultimate Theory of Human Parsing}
\title{Learning Compositional Neural Information Fusion for Human Parsing}

\author{Wenguan Wang\thanks{Equal contribution.} $~^{1,3}$, Zhijie Zhang\footnotemark[1] $~^{2,1}$, Siyuan Qi $~^{3}$, Jianbing Shen$~^{1}$, Yanwei Pang$~^{2}$\thanks{Corresponding author: \textit{Yanwei Pang}. }~, Ling Shao$~^{1}$\\
$^1$\small Inception Institute of Artificial Intelligence, UAE \\$^2$\small School of Electrical and Information Engineering, Tianjin University
~~~~$^3$\small University of California, Los Angeles, USA\\
{\tt\small \{wenguanwang.ai, teesoloj\}@gmail.com}\\
{\tt\small \url{https://github.com/ZzzjzzZ/CompositionalHumanParsing}}
}

\maketitle
\thispagestyle{empty}

%%%%%%%%% ABSTRACT
\begin{abstract}
%This work proposes to combine neural networks with structural hierarchy for efficient and complete human parsing. Essentially, the inference of our deep hierarchical human parser involves three processes: direct inference (directly using backbone network feature in each part), bottom-up inference (assembling information from constituent parts), and top-down inference (leveraging knowledge from parent nodes). Bottom-up and top-down inference explicitly model the composition and decomposition relations in human bodies.  More significantly, the executions of complicated bottom-up and top-down inference are conditioned on the inputs. These designs tie in with human visual perception process. In some {\color{red}easy cases} with clear and complete objects, we more rely on intuitive understanding (direct feature mapping). However, in some difficult scenes, a complicated inference is further needed, \ie, using bottom-up processing when seeing only parts of objects, and top-down, fine-grained processing for small objects. Such conditional inference strategy selectively activates the bottom-up/top-down process, reducing computational cost while maintaining modeling accuracy. Our approach is extensively evaluated on five popular datasets, showing that it establishes a new state-of-the-art in each case, with a fast processing speed of 23fps. Our code and results will be released to help ease future research in this direction.
This work proposes to combine neural networks with the compositional hierarchy of human bodies for efficient and complete human parsing. We formulate the approach as a neural information fusion framework. Our model assembles the information from three inference processes over the hierarchy: direct inference (directly predicting each part of a human body using image information), bottom-up inference (assembling knowledge from constituent parts), and top-down inference (leveraging context from parent nodes). The bottom-up and top-down inferences explicitly model the compositional and decompositional relations in human bodies, respectively.  In addition, the fusion of multi-source information is conditioned on the inputs, \ie, by estimating and considering the confidence of the sources. The whole model is end-to-end differentiable, explicitly modeling information flows and structures. %These designs tie in with human visual perception process. %In some {\color{red}easy cases} with clear and complete objects, we more rely on intuitive understanding (direct feature mapping). However, in some difficult scenes, a complicated inference is further needed, \ie, using bottom-up processing when seeing only parts of objects, and top-down, fine-grained processing for small objects. Such conditional inference strategy selectively activates the bottom-up/top-down process, reducing computational cost while maintaining modeling accuracy.
Our approach is extensively evaluated on four popular datasets, outperforming the state-of-the-arts in all cases, with a fast processing speed of 23fps. Our code and results have been released to help ease future research in this direction.
\end{abstract}

%%%%%%%%% BODY TEXT
\vspace{-5pt}
\section{Introduction}
\vspace{-3pt}
%Human parsing, which aims to decompose humans into semantic parts (\eg, arms, legs, \etc), is a crucial yet challenging task for detailed human body configuration analysis in 2D monocular images. It gained increasing attention for its essential role in many dependent application domains, such as surveillance analysis~\cite{ladicky2013human,gkioxari2015actions}, fashion synthesis~\cite{zhu2017your}, to name a few representative ones.

%Recent human parsing approaches have gained remarkable progress. The advances are made from two major aspects. 1) Improved deep neural network architectures for semantic segmentation: some representative works~\cite{chen2016attention,zhao2017self,Luo_2018_TGPnet} achieved the state-of-the-art performance with the help of well-designed architectures (\eg, fully convolutional networks (FCN)~\cite{long2015fully}, DeepLab~\cite{chen2018deeplab}, \etc). 2) Auxiliary information: some works tried to leverage extra human joints to better address human configurations~\cite{gong2017look,xia2017joint,nie2018mutual}, requiring additional training data of human keypoints.

%%%%%%%%%%%%%%%%%%% Figure 1 %%%%%%%%%%%%%%%%%%%%%%
\begin{figure}[t]
%%tr = 0.006, ts = 0.008
  \centering
      \includegraphics[width=0.99\linewidth]{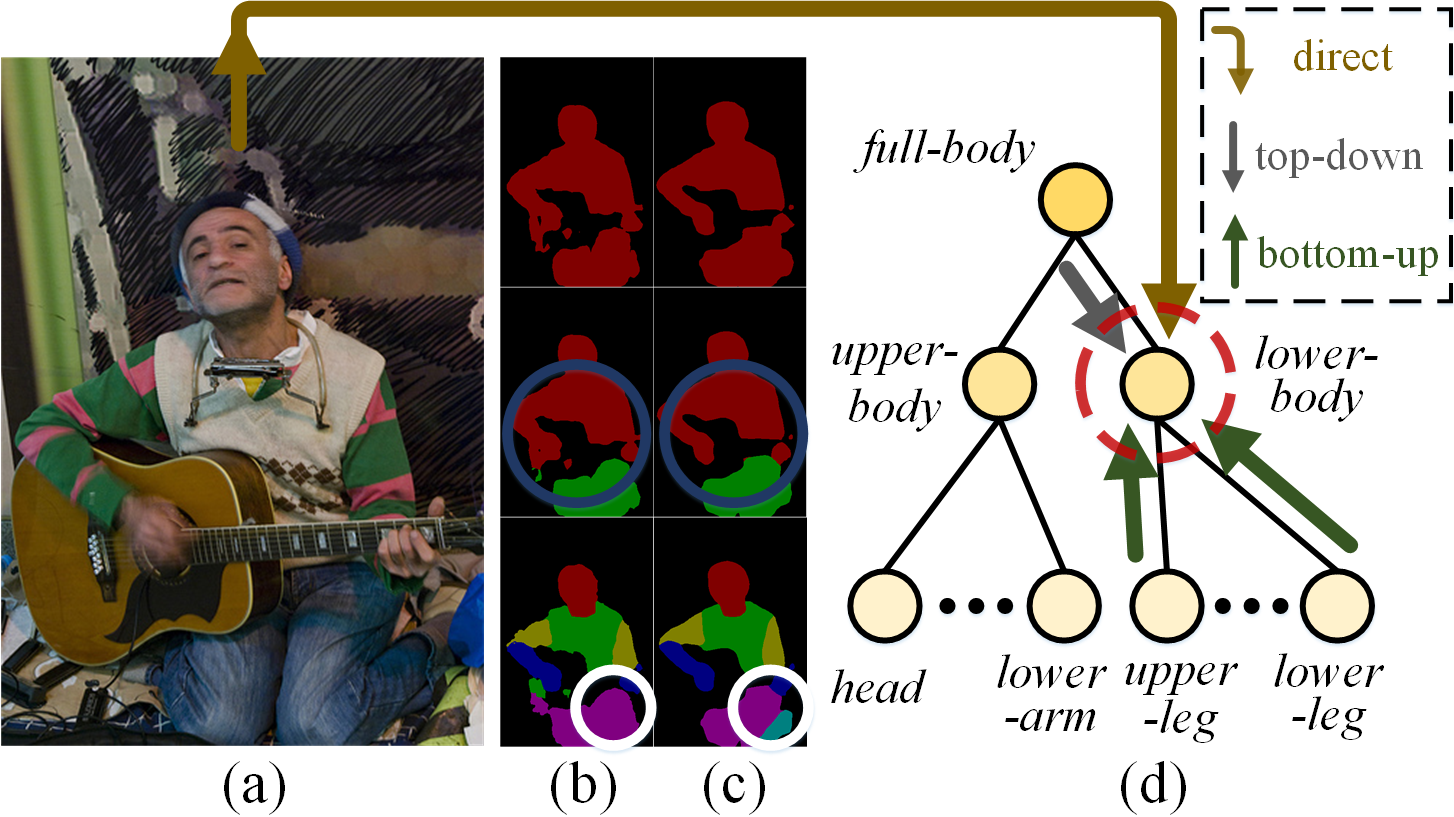}
\vspace{-10pt}
\caption{\small We represent the human body (a) as a hierarchy of multi-level semantic parts, and treat  human parsing as a multi-source information fusion process. For each part, information from three sources (direct, bottom-up, and top-down processes) are fused to better capture the structures in this problem. For clarity, we only show the information fusion of the \textit{lower-body} node in the red circle in (d). Compared to directly inferring the human semantics in (b), (c) shows better results after compositional neural information fusion (d).}
\label{fig:overview}
\vspace{-12pt}
\end{figure}

Human parsing, which aims to decompose humans into semantic parts (\eg, arms, legs, \etc), is a crucial yet challenging task for detailed human body configuration analysis in 2D monocular images. It has gained increasing attention owing to its essential role in many areas of application, such as surveillance analysis~\cite{ladicky2013human}, %,gkioxari2015actions
and fashion synthesis~\cite{zhu2017your}, to name a couple.

Recent human parsing approaches have made remarkable progress.
Some representative ones~\cite{chen2016attention,zhao2017self,Luo_2018_TGPnet} are built upon well-designed deep learning architectures for semantic segmentation (\eg, fully convolutional networks (FCNs)~\cite{long2015fully}, DeepLab~\cite{chen2018deeplab}, \etc). Though these achieve promising results, they fail to make full use of the rich structures in this task. %Although inspired by the rich structure in this task, they do not make full use of them.
Some others use extra human joints to better constrain body configurations~\cite{gong2017look,xia2017joint,nie2018mutual}, requiring additional training data of human keypoints and ignoring the compositional relations within human bodies.

In this paper, we segment body parts at multiple levels (see \autoref{fig:overview}), in contrast to most previous human parsers which only focus on atomic parts (represented as leaf nodes in the human hierarchy).
The insight is that estimating the whole graph provides us cross-level information that can assist learning and inference for each body part. This is also evidenced by human perception studies~\cite{kimchi1992primacy,tarr1998image,navon1977forest}; a global shape can either precede or follow the recognition of its local parts, and both contribute to the final recognition.

We further specify this as a \textit{multi-source information fusion} procedure, which integrates information from the following three processes  (see \autoref{fig:overview} (d)). \textbf{1)} \textit{Direct inference} (or unconscious inference) from the input image. For example, sometimes humans directly recognize objects ny relying on intuitive understanding~\cite{marcel1983conscious,wu2011numerical}. %When suffering more difficult scenes, humans may seek to more complex analysis.
\textbf{2)} \textit{Top-down inference}, which recognizes
fine-grained components from a whole entity. For example, when recognizing small fine-grained parts, exploring contextual information of the entire object is essential~\cite{gregory1970intelligent,grill2014functional,wang2019iterative}  (see the regions in the white circles in \autoref{fig:overview}). \textbf{3)} \textit{Bottom-up inference}, which associates constituent parts to predict upper-level nodes. When objects are partially occluded or contain complex topologies, humans can assemble sub-parts to assist in recognizing the entities~\cite{epshtein2008image,gibson1966senses} (see the regions in the blue circles in \autoref{fig:overview}).

Employing the strong learning power of deep neural networks~\cite{hornik1989multilayer,LiangS17}, we build a compositional neural information fusion for these three inference processes in an end-to-end manner. This yields a hierarchical human parsing framework to better capture the compositional constraints and human part semantics. In addition, we design our model as a conditional fusion, \ie, the assembly of different information is dependent on the confidence estimations for the sources, instead of simply assuming all the sources are reliable. This is achieved by a learnable gate mechanism, leading to more accurate parsing results.

This paper makes three contributions. 1) We formulate the human parsing problem as a neural information fusion process over a compositionally structured network. 2) We analyze three important sources of information, leading to a novel network architecture that conditionally incorporates direct, top-down, and bottom-up inferences. 3) Our model achieves state-of-the-art performances for comprehensive evaluations on four public datasets (LIP~\cite{gong2017look}, PASCAL-Person-Part~\cite{xia2017joint}, ATR~\cite{liang2015deep} and Fashion Clothing~\cite{Luo_2018_TGPnet}). Testing with more than 20K images demonstrates the superiority over existing methods of exploiting compositional structural information for human parsing. %It also achieves a score of 60.51\% mIOU on LIP test set at the time of submission, which surpasses the winning entry of LIP Challenge 2018 (57.90\% mIOU).

\vspace{-3pt}
\section{Related Work}
\vspace{-3pt}
% Our study is closely related to the following fields: semantic human part parsing, adaptive computation for neural networks, and hierarchical models in computer vision.

\noindent\textbf{Hierarchical/Graphical Models in Computer Vision:} Hierarchical/graphical models are powerful for building structured representations, which can reflect task-specific relations and constraints. From early distributional semantic models, part-based models~\cite{felzenszwalb2010object,felzenszwalb2005pictorial}, MRF/CRF~\cite{joachims2009cutting}, And-Or grammar model~\cite{AOG2009object}, to deep structural networks~\cite{jain2016structural,fang2018learning}, graph neural networks~\cite{gilmer2017neural}, trainable CRF~\cite{zheng2015conditional}, \etc, hierarchical/graphical models have found applications in a wide variety of core computer vision tasks, such as object recognition~\cite{qi2018learning}, human parsing~\cite{liang2016semantic,liang2016semantic2,Zhu2018ProgressiveCH}, pose estimation~\cite{ladicky2013human,wang2015joint,tang2018deeply,wang2018attentive,li2019crowdpose}, visual dialog \etc,
to the extent that they are now ubiquitous in the field. Inspired by their general success, we leverage structural information to design our approach.
% Inspired by their general success, our method augments hierarchical human semantic representations with the learning capability of neural networks.
In addition to directly inferring segments from the image features, we further derive two additional inference processes, \ie, bottom-up and top-down inference, to better capture human structures. This encourages more reasonable results that are consistent with the human body configuration.

\noindent \textbf{Information Fusion:} Our method is also inspired by the idea of fusing information from different sources to obtain a better prediction of the target. One typical application of this is sensor fusion, which is a broad field, discussed in more detail in~\cite{khaleghi2013multisensor}.
%which is a broad field that we refer the readers to~\cite{khaleghi2013multisensor} for a thorough treatment.
Many machine learning models can be regarded as information fusion methods: \eg, product of experts~\cite{hinton1999products}, Bayesian fusion, ensemble methods~\cite{dietterich2000ensemble}, and graphical models~\cite{wainwright2008graphical}. Motivated by this general idea, we learn to adaptively fuse the direct inference along with top-down and bottom-up predictions in the compositional human structure for our final prediction.

\noindent\textbf{Human Parsing Models:} Traditional human parsing models are typically built upon hand-crafted visual features (\eg, color, HoG)~\cite{yamaguchi2013paper,liu2014fashion,wang2011blocks,yamaguchi2012parsing,luo2013pedestrian,simo2014high,yang2014clothing}, low-level image decompositions (\eg, super-pixel)~\cite{liu2014fashion,yamaguchi2012parsing,yang2014clothing}, and heuristic hypotheses (\eg, grammars for human body configuration)~\cite{chen2006composite,dong2013deformable,chen2014detect,dong2014towards}. Though impressive results have been achieved, these pioneering works require a lot of carefully hand-designed pipelines, and suffer the limited representability of the hand-crafted features.

With the renaissance of connectionism in the computer vision community, recent research efforts take deep neural networks as their main building blocks~\cite{xia2016zoom,nie2018mutual,zhao2017self,luo2018macro,Luo_2018_TGPnet,zhao2018understanding,liu2018cross}. More specifically, some efforts address the task as an active template regression problem~\cite{liang2015deep}, propagate semantic information from a retrieved, annotated image corpus~\cite{liu2015matching}, merge multi-level image context in a unified convolutional neural network~\cite{liang2015human}, or use Graph LSTMs to model human configurations~\cite{liang2016semantic,liang2016semantic2}. Some others leverage extra pose information to assist the task~\cite{xia2016pose,gong2017look,xia2017joint,fang2018weakly,nie2018mutual}.
%\ie, use a structure-sensitive learning to enforce the semantic consistency between parsing results and human joint structures~\cite{gong2017look}, or  mutual nie2018mutual,.
In contrast to the above approaches addressing category-level understanding of human semantics, a few methods operate at an instance level~\cite{li2017holistic,zhou2018adaptive,CE2P2019}.%gong2018instance,

The aforementioned deep human parsers generally achieve promising results, due to the strong learning power of neural networks~\cite{long2015fully,chen2018deeplab} and the plentiful availability of  annotated data~\cite{gong2017look,xia2017joint}. However, they typically need to pre-segment images into superpixels~\cite{liang2016semantic,liang2016semantic2}, which breaks the end-to-end story and is time-consuming, or rely on extra human landmarks~\cite{xia2016pose,gong2017look,xia2017joint,fang2018weakly,nie2018mutual}, requiring additional annotations or pre-trained pose estimators. Though~\cite{Zhu2018ProgressiveCH} also performs multi-level, fine-grained parsing, it neither explores different information flows within human hierarchies nor models the problem from the view of multi-source information fusion.

In contrast, we elaborately design a compositional neural information fusion framework, which explicitly captures human compositional structures and dynamically combines direct, bottom-up and top-down inference modes over the hierarchy. The overall model inherits the complementary advantages of FCNs and hierarchical models, yielding a unified, end-to-end trainable human parsing framework with a strong learning ability, improved representational power, as well as high processing speed.

%\noindent\textbf{Adaptive Computation for Neural Networks:} To seek a better trade-off between network depth and computation cost, some works~\cite{bengio2013estimating,bengio2015conditional} explored to dynamically activate different parts of a network in an input-dependent fashion.
%Some researchers~\cite{bengio2015conditional,liu2018dynamic,wu2018blockdrop} viewed this  as a sequential decision making problem, \ie, at each layer, given current state of the computation, deciding which units to activate. They therefore use reinforcement learning to learn conditional computation policies. Some more recent methods were proposed to selectively allocate computation of Residual Network~\cite{he2016deep} per image region~\cite{figurnov2017spatially}, or equip each residual module with a learnable gate (determines whether to execute or skip the layer)~\cite{veit2018convolutional}. In this work, for the first time, we introduce adaptive computation for human parsing. Through conditioning the executions of complicated bottom-up and top-down inference on the inputs, we gain an on-the-fly inference topology with faster speed while remaining the excellent performance.

%Our model is generally fall in this category.
\vspace{-3pt}
\section{Our Approach}
\vspace{-3pt}
% We solve a slightly augmented problem comparing to the typical human parsing problem, which often refers to segmenting a human image into fine-grained semantic parts (leaf nodes in the human structural hierarchy). Here we find the segmentation of body parts of all levels. The insight is that estimating the whole graph gives us cross-level information that can assist inference for each body part. For example, a detected upper body can help locate the left arm (top-down) and the full body (bottom-up). Please be noted that such heuristic setting would not bring any extra annotation effort in the segmentation problem.

\begin{figure}[t]
%%tr = 0.006, ts = 0.008
  \centering
      \includegraphics[width=0.99 \linewidth]{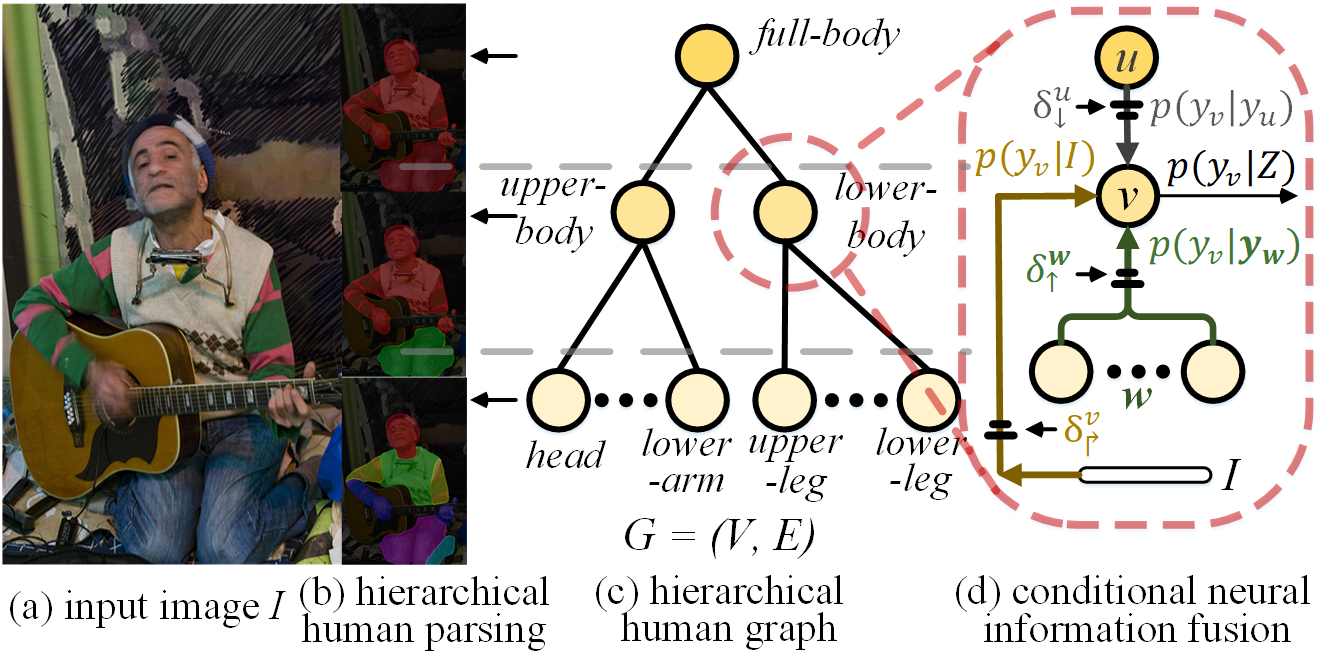}
\vspace{-9pt}
\caption{\small Given an input image (a), our compositional and conditional neural information fusion is performed over the human graph (c) to produce hierarchical parsing results.}
\label{fig:compositional_fusion}
\vspace{-12pt}
\end{figure}

Formally, we represent the hierarchical human body structure as a graph $\mathcal{G}\!=\!(\mathcal{V}, \mathcal{E}, \mathcal{Y})$, where nodes $v\!\in\!\mathcal{V}$ represent human parts in different levels, and edges $e\!\in\!\mathcal{E}$ are two-tuples $e\!=\!(u, v)$  representing the compositional relation that node $v$ is a part of node $u$. As shown in \autoref{fig:compositional_fusion} (c), the nodes are further grouped into $L(=\!3)$ levels: $\mathcal{V}\!=\!\mathcal{V}^1\!\cup\!\dots\!\cup\!\mathcal{V}^L$, where $\mathcal{V}^1$ are the leaf nodes (the most fine-grained semantic parts typically considered in common human parsers), $\mathcal{V}^{2\!}\!=$\{\textit{upper-body, lower-body}\}, and $\mathcal{V}^{3\!}\!=$\{\textit{full-body}\}. For each node $v$, we want to infer a segmentation map $y_{v}\!\in\!\mathcal{Y}$ that is a probability map of its label.  Please note that such a problem setting does not introduce any additional annotation requirement, since higher-level annotations can be obtained by simply combining the lower-level labels.

There are three different sources of information when inferring $y_{v}$ for $v$: 1) the raw input image, 2) $y_{u}$ for the parent node $u$, and 3) $\bm{y_{w}}$ for all the child nodes $\bm{w}$.
We treat the final prediction of $y_v$ as a fusion of the information from these three sources. Next, we briefly review different methods to modeling this information fusion problem that motivate our solution and network design for human parsing.

\vspace{-3pt}
\subsection{Information Fusion}
\vspace{-3pt}
Information fusion refers to the process of combining information from several sources $Z\!=\!\{z_1, z_2, \cdots, z_n \}$ in order to form a unified picture of the measured/predicted target $y$. Each source provides an estimation of the target. These sources can be the raw data $x$ or some other quantities that can be inferred from $x$. Several approaches have been proposed to tackle this problem.

\noindent$\bullet$ Product of experts (PoE)~\cite{hinton1999products} treats each source as an ``expert". It multiplies the probabilities and then renormalizes:
\vspace{-5pt}
\begin{equation}\small
\begin{aligned}
p(y|Z) = \frac{\prod_{i=1}^{n} p(y|z_i)}{\sum_{y} \prod_{i=1}^{n} p(y|z_i)}.
\end{aligned}
%\vspace{2pt}
\end{equation}
% The gradient for the PoE is usually computationally infeasible. Hence instead of directly optimizing the data likelihood, learning the PoE is often achieved by optimizing the contrastive divergence~\cite{hinton2002training} as an alternative objective. The main reason is that the gradient of contrastive divergence is easier to approximate than MLE.

\noindent$\bullet$ Bayesian fusion. Denoting $Z_{s\!}\!=\!\{z_1, z_2, \!\cdots\!, z_s\}$ as the set of the first $s$ sources, it factorizes the posterior probability:
\vspace{-0pt}
\begin{equation}\small
\begin{aligned}
\!\!\!\!p(y|Z)\!=\!\frac{p(Z_n|y)p(y)}{p(Z_n)}\!=\!\frac{p(y)p(z_1|y)\prod_{s=2}^{n} p(Z_{s}|Z_{s-1}, y)}{p(z_1)\prod_{s=2}^{n} p(Z_{s}|Z_{s-1})}.\!\!
\end{aligned}
\vspace{-0pt}
\end{equation}
However, it is too difficult to learn all the conditional distributions. By assuming the independence of different information sources, we have the Naive Bayes:
\vspace{-4pt}
\begin{equation}\small
\begin{aligned}
p(y|Z) \propto p(y)\prod\nolimits_{i} p(z_i|y),
\end{aligned}
\vspace{-4pt}
\end{equation}
which serves as an approximation of the true distribution.

\noindent$\bullet$ Ensemble methods. In this approach, each $z_i$ is a classifier that predicts $y$. A typical ensemble method is Bayesian voting~\cite{dietterich2000ensemble}, which weights the prediction of each classifier to get the final prediction:
\vspace{-5pt}
\begin{equation}\small
\begin{aligned}
p(y|Z)  = \sum\nolimits_{z_i} p(y|z_i) p(z_i|x).
\end{aligned}
\vspace{-5pt}
\end{equation}
The AdaBoost~\cite{freund1997decision} algorithm also falls into this category.

\noindent$\bullet$ Graphical models (\eg, conditional random fields). In such models, each $z_i$ can be viewed as a node that contributes to the conditional probability:
\vspace{-5pt}
\begin{equation}\small
\begin{aligned}
p_{\theta}(y|Z) = \exp \{ \sum\nolimits_{i}\phi_{\theta_i}(y, z_i) - A(\theta) \},
\end{aligned}
\vspace{-5pt}
\end{equation}
where $A(\theta)$ is the log-partition function that normalizes the distribution. Computing $A(\theta)$ is often intractable, hence the solution is usually given by approximation methods, such as Monte Carlo methods or (loopy) belief propagation~\cite{sutton2012introduction}.

\vspace{-3pt}
\subsection{Compositional Neural Information Fusion}
\vspace{-3pt}
The above methods can all be viewed as ways to approximate the true underlying distribution $p(y|Z)$, which can be written as a function of predictions from different information sources $Z$:
\vspace{-5pt}
\begin{equation}\small
\begin{aligned}
p(y|Z) = f(p(y|z_1), p(y|z_2), \cdots, p(y|z_n)).
\end{aligned}
\vspace{-5pt}
\end{equation}
There are potential drawbacks to following the exact solution of one of the above methods. First, they are not entirely consistent with each other. For example, the PoE multiplies all $p(y|z_i)$ together, whereas ensemble methods compute their weighted sum. Each method approximates the true distribution in a different way and has its own tradeoff. Second, exact inference is difficult and solutions are often approximative (\eg, contrastive divergence~\cite{hinton2002training} is used for PoE and Monte Carlo methods for graphical models).
%%%%%%%%%%%%%%%%%%% Figure 2%%%%%%%%%%%%%%%%%%%%%%
\begin{figure*}[!th]
%%tr = 0.006, ts = 0.008
  \centering
      \includegraphics[width=1 \linewidth]{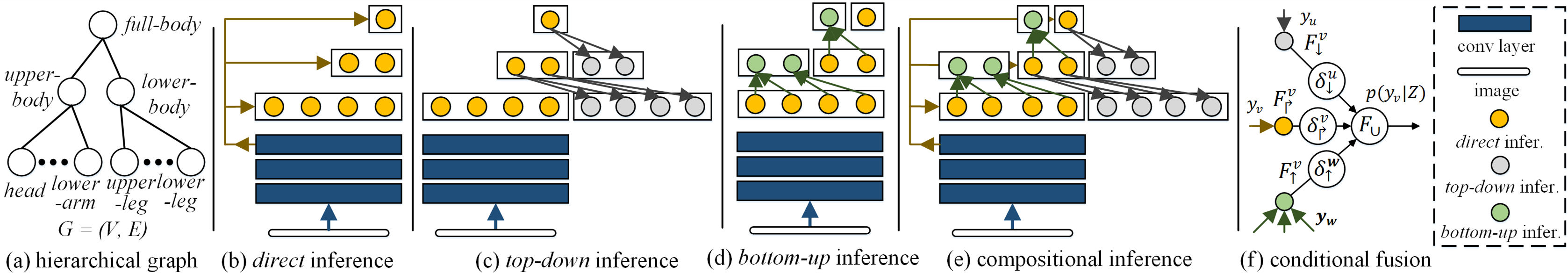}
\vspace{-20pt}
\caption{\small Illustration of our conditional neural information fusion network for hierarchical human parsing. See text for details.}
\label{fig:model}
\vspace{-12pt}
\end{figure*}

Therefore, instead of exactly following the computation of one of the above methods, we leverage neural networks to directly model this fusion function, due to their strong ability for flexible feature learning and function approximation~\cite{hornik1989multilayer,LiangS17}. The hope is that we can directly learn to fuse multi-source information for a specific task.

However, the fusion network should not be learned arbitrarily without inductive biases~\cite{craik1967nature,mitchell1980need,battaglia2018relational}, which is the preference for structural explanations exhibited in human reasoning processes. Here, we exploit the compositional nature of the problem and design the network with the following observations:

\noindent $\bullet$ In the compositional structure $\mathcal{G}$, the final prediction $p(y_v|Z)$ for each node $v$ combines information from three different sources: 1) the direct inference $p(y_v|x)$ from the raw image input, 2) the top-down inference $p(y_v|y_u)$ from the parent node $u$, which utilizes the \textbf{decompositional} relation, and 3) the bottom-up inference $p(y_v|\bm{y_w})$, which assembles predictions $\bm{y_w}$ for all the child nodes $\bm{w}$ to leverage the \textbf{compositional} relation.

\noindent $\bullet$ In many cases, simply fusing different estimations could be problematic. The final decision should be conditioned on the \textbf{confidence} of each information source.

Based on the above observations, we design our parser network to learn a \textit{compositional neural information fusion}:
\vspace{-7pt}
\begin{equation}\small
\begin{aligned}
p(y_v|Z) &\!=\!f(\,\delta^v_{\Rsh}p(y_v|x),\, \delta^u_{\downarrow}p(y_v|y_u),\, \delta^{\bm{w}}_{\uparrow}p(y_v|\bm{y_w})\,),
\end{aligned}
\vspace{0pt}
\end{equation}
where the confidence $\delta$ is a learnable continuous function with outputs from 0 to 1. The symbols $\Rsh$, $\downarrow$, and $\uparrow$ denote direct, top-down, and bottom-up inference, respectively. As shown in \autoref{fig:compositional_fusion} (d), this function fuses information from the three sources in the compositional structure, taking into account the confidence of each source. For neural network realizations of this function, the probability terms can be relaxed to logits, which are essentially log-probabilities.

When carrying out such a prediction, there is one computational issue.
Notice that the top-down/bottom-up inferences rely on an estimation of the parent/child node(s). This forms a circular dependency between a parent and its children. To solve this, we treat the direct inference result from the raw data as an initial estimation, and the top-down/bottom-up inferences rely on this initial estimation\footnote{For some nodes, bottom-up or top-down inference might not exist. The terminal leaf nodes $\mathcal{V}^1$ do not have bottom-up inference, while the root node $\mathcal{V}^3$
only has direct and bottom-up inference. For clarity of the method description, we discuss the general case with all three sources.}. Therefore, we decompose the algorithm into three consecutive steps:

\noindent\textbf{1.}~\textbf{Direct inference.} Given the raw data as input, we assign an estimation $\tilde{y}_v$ for each node $v\!\in\!\mathcal{V}$.

\noindent\textbf{2.}~\textbf{Top-down/bottom-up inference.} We estimate $p(y_v|\tilde{y}_u)$ and $p(y_v|\tilde{\bm{y}}_{\bm{w}}))$ based on the estimated $\tilde{y}_u$ and $\tilde{\bm{y}}_{\bm{w}}$ in step 1.

\noindent\textbf{3.}~\textbf{Conditional information fusion.} Based on the above results, we obtain a final prediction for each node $v$ by $y_v^*\!=\!\argmax_y f(\delta^v_{\Rsh}p(y_v|x),\, \delta^u_{\downarrow}p(y_v|\tilde{y}_u),\, \delta^{\bm{w}}_{\uparrow}p(y_v|\tilde{\bm{y}}_{\bm{w}}))$.
% \begin{equation}\small
% \begin{aligned}
% \!\!y_v^*\!=\!\argmax_y f(p(y_v|x),\, \delta^u_{\downarrow}p(y_v|\tilde{y_u}),\, \delta^{\bm{w}}_{\uparrow}p(y_v|\tilde{\bm{y}}_{\bm{w}})).\!\!
% \end{aligned}
% \end{equation}

This procedure motivates the overall network architecture, where each step above can be learned as a module by a neural network. Next, we discuss our network design.
%In the next section, we discuss the details of our neural network implementation for human parsing.

\vspace{-3pt}
%-------------------------------------------------------------------------
\subsection{Network Architecture}
\vspace{-3pt}
%Based on the above algorithm, the complete network architecture stacks the following three parts to form an end-to-end system for human parsing:
%
%\textbf{Direct inference network.} Given the raw input image $I$, the direct inference network predicts a segmentation map $\tilde{y_v}$ for each node $v \in \mathcal{V}$ that represents a human part.
%
%Global feature (backbone): $$h_{\mathcal{V}} = F_g (I; W_g)$$
%Node feature: $$h_v = F_v (h_{\mathcal{V}})$$
%$$p(\tilde{y_v} | x) = F_d (h_v; W_v) $$

%Here we describe the overall architecture based on the above algorithm. The detailed network will be introduced in \autoref{sec:network}.
Our model stacks the following parts to form an end-to-end system for hierarchical human parsing. The system does not require any preprocessing and the modules are FCNs, so it achieves high efficiency.

\noindent\textbf{Direct Inference Network.} This directly predicts a segmentation map $\tilde{y}_v$ for each node $v$ (a human part), using information from the image (see \autoref{fig:model} (b)). Formally, given an input image $I\!\in\! \mathbb{R}^{K\times K\times 3}$, a backbone network $B$ (\ie, a DeepLabV3-like network, parameterized by $\mathbf{W}_{\!\!B}$) is first employed to obtain a new effective image representation $h_{I}$:
\begin{equation}\small
\vspace{-4pt}
\begin{aligned}
\text{\small image embedding:}~~~~~h_{I}=F_{\text{B}}(I; \mathbf{W}_{\!\!B})\!\in\! \mathbb{R}^{k\times k\times c}.
\end{aligned}
\label{eq:imageembedding}
\vspace{-0pt}
\end{equation}
As the nodes $\mathcal{V}$ capture explicit semantics, a specific feature ${h}_v$ for each node $v$ is desired for more efficient representation. However,
%Then, for each node $v$, a FCN-based embedding function $E^v$ is used to generate its \textit{node embedding/feature} $h_v\!\in\! \mathbb{R}^{w\times h\times c}$ from the image feature $h_{I}$.
%\begin{equation}\small
%\begin{aligned}
%\text{\small node embedding:}~~~~~{h}_v &= E^v ({h}_{I}; \mathbf{W}^v_E).
%\end{aligned}
%\label{eq:nodeembedding}
%\end{equation}
%To achieve $E^v$, designing a node-specific FCN will lead to heavy computational cost, while a single weight-sharing model may loose node-specific semantics. To address this, for each node, we first use a \textit{level-specific FCN} (LSFCN), which shares weights across the same-level nodes (\eg, \textit{upper-body}, \textit{lower-body}), to capture the semantics in different levels and relations between the same-level nodes:
using several different, node-specific embedding networks will lead to a high computational cost. To remedy this, for each $l$-th level, we first apply a \textit{level-specific} FCN (LSF) to describe the level-wise semantics and contextual relations:
\begin{equation}\small
\begin{aligned}
\!\!\!\!\text{\small level-specific embedding:}~~{h}^l_{\text{LSF}} \!=\!F^l_{\text{LSF}} ({h}_{I}; \mathbf{W}^l_{\!{\text{LSF}}})\!\in\! \mathbb{R}^{k\times k\times c\!\!},\!\!
\label{eq:levelembedding}
\end{aligned}
\end{equation}
where $l\!\in\!\{1, 2, 3\}$. More specifically, three LSFs (${F}^1_{\text{LSF}}$, ${F}^2_{\text{LSF}}$, and ${F}^3_{\text{LSF}}$) are learned to extract three level-specific embeddings (${h}^1_{\text{LSF}}, {h}^2_{\text{LSF}}$, and ${h}^3_{\text{LSF}}$).
Further, for each node $v$, an independent channel-attention block, Squeeze-and-Excitation (SE)~\cite{senet}, is applied to obtain its specific feature:
\begin{equation}\small
\begin{aligned}
\!\!\text{\small node-specific embedding:}~~{h}_v \!=\!F^v_{\text{SE}}({h}^{l}_{\text{LSF}}; \mathbf{W}^v_{\!{\text{SE}}})\!\in\! \mathbb{R}^{k\times k\times c\!\!},\!\!
\end{aligned}
\label{eq:seembedding}
\end{equation}
where $v\!\in\!\mathcal{V}^{l}$ (\ie, $v$ is located in the $l$-th level). By explicitly modelling the interdependencies between channels, $F^v_{\text{SE}}$ allows us to adaptively recalibrate the channel-wise features of ${h}^{l}_{\text{LSF}}$ to generate node-wise representations. Meanwhile, due to its light-weight nature, we can achieve our goal with minimal computational overhead.
Then, the direct inference network ${F}_{\Rsh}$ reads the feature and predicts the segmentation map $\tilde{y}_v$:
\vspace{-3pt}
\begin{equation}\small
\begin{aligned}
\text{logit}(\tilde{y}_v | I)= {F}_{\Rsh} (h_v; \mathbf{W}_{\Rsh})\!\in\! \mathbb{R}_{\geq 0}^{k\times k}.
\end{aligned}
\vspace{-3pt}
\label{eq:flsh}
\end{equation}
%\textbf{Top-down/bottom-up network.} Based on the outputs from the direct inference network, the top-down/bottom-up network predicts the segmentation maps based on the compositional structure of human bodies. Specifically, the top-down network takes as input the initial estimation $\tilde{y_v}$ of a non-leaf node $v$, and predicts the segmentation maps of all its child nodes. The bottom-up network works the opposite way; it predicts the segmentation of a non-leaf node and take the estimation of all its child nodes as input.
%
%$$p(y_v | y_u) = F_{\downarrow} (y_v | y_u; h_v, W_{v\downarrow})$$
%$$p(y_v | y_{\bm{w}}) = F_{\uparrow} (y_v | y_{\bm{w}}; h_v, W_{\bm{w}\uparrow})$$

\noindent\textbf{Top-down Inference Network.}
Based on the outputs from the direct inference network, the top-down inference predicts segmentation maps by considering human decompositional structures. Specifically, for node $v$, the top-down network $F_{\downarrow}$ leverages the initial estimation $\tilde{y}_u$ of its parent node $u$ as high-level contextual information for prediction (see \autoref{fig:model} (c)):
\vspace{-4pt}
\begin{equation}\small
\begin{aligned}
\!\!\text{logit}(y_v | \tilde{y}_u)\!=\!F_{\downarrow} (y_v | \tilde{y}_u; h_v, \!\!\mathbf{W}_{\downarrow})\!=\!F_{\downarrow}([\tilde{y}_u, h_v]) \in \mathbb{R}_{\geq 0}^{k\times k}.\!\!
\end{aligned}
\label{eq:top-down}
\end{equation}
Here, the concatenated feature $[\tilde{y}_u, h_{v}]$ is fed into the FCN-based $F_{\downarrow}$, parameterized by $\mathbf{W}_{\downarrow}$, for top-down inference.

\noindent\textbf{Bottom-up Inference Network.} One major difference to the top-down network is that, for each node $v$, the bottom-up network needs to gather information (\ie, $\tilde{\bm{y}}_{\bm{w}}\!\in\!\mathbb{R}_{\geq 0}^{k\times k\times |\bm{w}|}$) from multiple descendants $\bm{w}$. Thanks to the compositional relations between $\bm{w}$ and $v$, we can transform $\tilde{\bm{y}}_{\bm{w}}$ to a fixed one-channel representation $\tilde{y}_{\bm{w}}$ through \textit{position-wise max-pooling} PMP (across channels):
\vspace{-2pt}
\begin{equation}\small
\begin{aligned}
\tilde{y}_{\bm{w}} = \text{PMP}([\tilde{y}_{w}]_{w\in\bm{w}})\in\mathbb{R}_{\geq 0}^{k\times k\times 1},
\end{aligned}
\label{eq:bottom-up1}
\vspace{-2pt}
\end{equation}
where $[$$\cdot$$]$ is a concatenation operation. Then, the bottom-up network $F_{\uparrow}$ gives a prediction according to compositional relations (see \autoref{fig:model} (d)):
\vspace{-2pt}
\begin{equation}\small
\begin{aligned}
\!\!\!\!\text{logit}(y_v | \tilde{\bm{y}}_{\bm{w}})\!=\!F_{\uparrow} (y_v | \tilde{\bm{y}}_{\bm{w}}; {h}_v,\!\!\mathbf{W}_{\uparrow})\!=\!F_{\uparrow} ([\tilde{y}_{\bm{w}}, {h}_v])\in \mathbb{R}_{\geq 0}^{k\times k\!}.\!\!
\end{aligned}
\vspace{-2pt}
\label{eq:bottom-up2}
\end{equation}

\noindent\textbf{Conditional Fusion Network.}
Before making the final prediction, we estimate the confidence $\delta$ of each information source using a neural gate function. For the direct inference of a node $v$, we estimate the confidence by:
\vspace{-2pt}
\begin{equation}\small
\begin{aligned}
\delta^v_{\Rsh} = \sigma(\mathbf{C}^v_{\Rsh} \cdot \text{CAP}(h_v))\in [0,1],
\end{aligned}
\vspace{-2pt}
\end{equation}
where $\sigma$ is the \textit{sigmoid} function. Here, CAP stands for \textit{channel-wise average pooling}, which has been proved a simple yet effective way for capturing the global statistics of convolutional features~\cite{senet,veit2018convolutional}. $\mathbf{C}^u_\Rsh\!\in\!\mathbb{R}^{1\times C}$ indicates a small fully connected layer that maps the $C$-dimensional statistic vector $\text{CAP}(h_v)\!\in\!\mathbb{R}^{C}$ of $h_v$ into a confidence score.

The confidence scores for the top-down and bottom-up processes follow a similar computational framework:
\vspace{-2pt}
\begin{equation}\small
\begin{aligned}
\delta^u_{\downarrow} &= \sigma(\mathbf{C}^u_{\downarrow} \cdot \text{CAP}(h_u))\in [0,1],\\
\delta^{\bm{w}}_\uparrow &= \sigma(\mathbf{C}^{\bm{w}}_\uparrow \cdot \text{CAP}([h_w]_{w\in{\bm{w}}}))\in [0,1],\\
\end{aligned}
\vspace{-2pt}
\end{equation}
where $\mathbf{C}^{u\!\!}_{\downarrow\!}\!\in_{\!}\!\!\mathbb{R}^{1\!~\times\!~C\!\!}$ and $\mathbf{C}^{\bm{w}\!\!}_\uparrow\!\in\!\mathbb{R}^{1\!~\times \!~C|\bm{w}|\!}$. Specificially, for the bottom-up process, we concatenate all the child node embeddings $[h_{w}]_{w\in\bm{w}\!}\!\in\!\mathbb{R}^{k\times k \times C|\bm{w}|\!}$. This means our decision is made upon the confidence of the union of the child nodes. Here, the confidence of a source can be viewed as a global score or statistic for interpreting the quality of the feature, which is learnt in an implicit manner.

\begin{figure}[t]
%%tr = 0.006, ts = 0.008
  \centering
      \includegraphics[width=0.99 \linewidth]{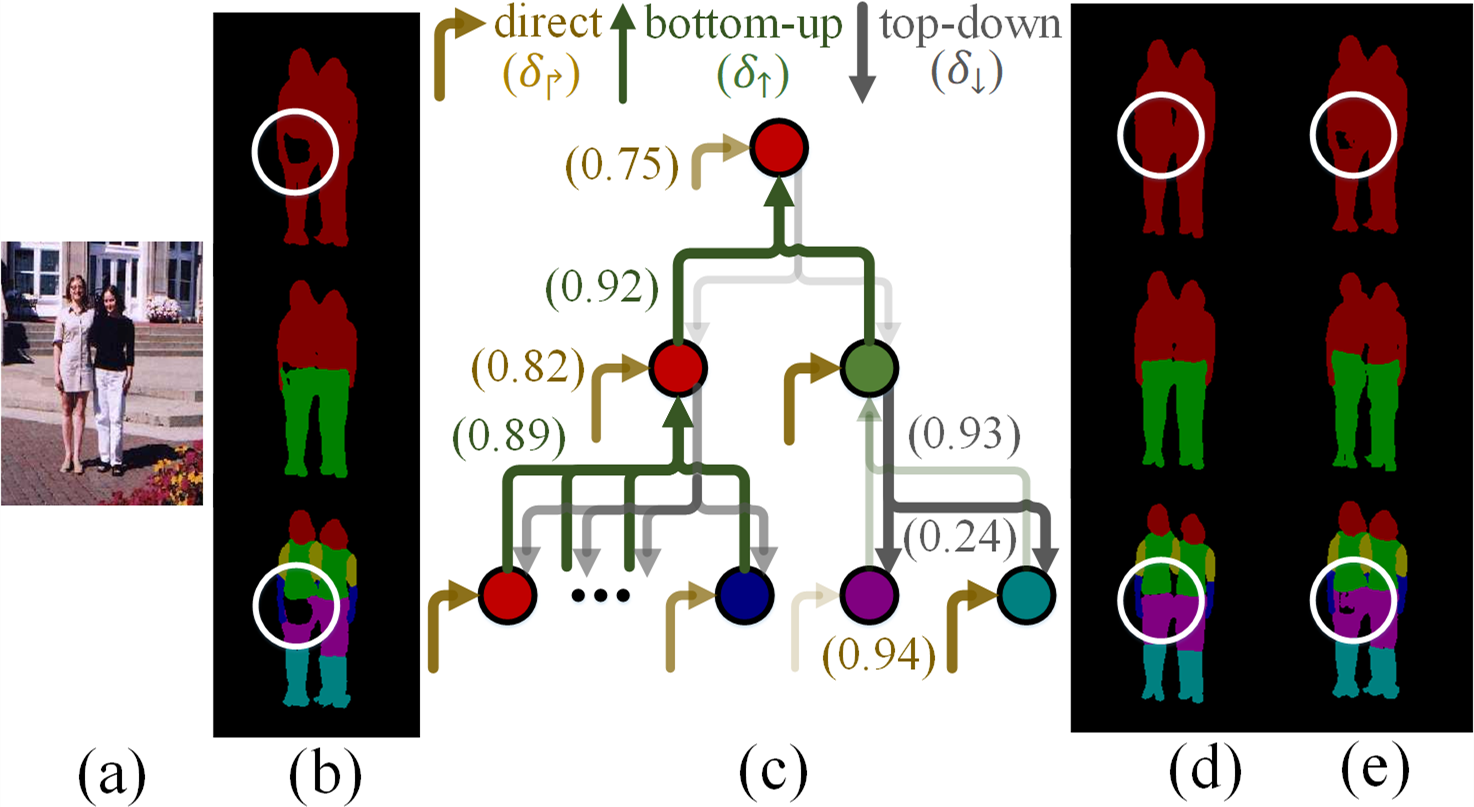}
\vspace{-9pt}
\caption{\small Illustration of our compositional inference and conditional fusion. (a) Input image. (b) Parsing results of direct inference. (c) Conditional information fusion, where the arrows with darker colors indicate higher values of gates $\delta$. For clarity, in ($\cdot$) we only show the gate values for a few inference processes. (d) Parsing results w/ compositional inference and conditional fusion. (e) Parsing results of compositional inference only. The improved regions are highlighted in white circles.}
\label{fig:dynamicinference}
\vspace{-14pt}
\end{figure}

%\noindent\textbf{Information fusion network.}
Finally, for each node $v$, the fusion network $F_{\cup}$ combines the results from the three inference networks above for final prediction (see \autoref{fig:model} (e)):
\vspace{-3pt}
\begin{equation}\small
\begin{aligned}
\text{logit}(y_v | Z)\!=\!F_{\cup}(\delta^v_{\Rsh}{F}^v_{\Rsh}, \delta^u_{\downarrow}F^v_{\downarrow}, \delta^{\bm{w}}_{\uparrow}F^v_{\uparrow}; \mathbf{W}_{\cup})\in \mathbb{R}_{\geq 0}^{k\times k \times 1\!},
\end{aligned}
\label{eq:fusion}
\vspace{-3pt}
\end{equation}
where $F_{\cup}\!:\!\mathbb{R}_{\geq 0}^{k\times k\times3}\!\rightarrow\!\mathbb{R}_{\geq 0}^{k\times k\times1}$ is implemented by a small FCN, parameterized by $\mathbf{W}_{\cup}$. \autoref{fig:dynamicinference} provides an illustration of our conditional fusion process. As can be seen, $\delta$ provides a learnable gate mechanism that suggests how much information can be used from a source (see \autoref{fig:dynamicinference} (c)). It is able to dynamically change the amount of information for different inference processes, \ie, condition on the sources (see \autoref{fig:dynamicinference} (d)). Thus, it yields better results than statically fusing the information with a weight-fixed fusion function (see \autoref{fig:dynamicinference} (e)). More detailed studies of our conditional and compositional fusion can be found in \autoref{sec:ablation} and \autoref{fig:abs}.

\noindent\textbf{Loss Function.} To obtain the final segmentation map from $\text{logit}(y_v|Z)$, we apply a \textit{softmax} function over the logits of nodes in the same level. Thus, for each level, all the inference networks; $F_{\Rsh}, F_{\downarrow}, F_{\uparrow}$, and the fusion network $F_{\cup}$ are trained by the standard cross-entropy loss:
\vspace{-3pt}
\begin{equation}\small
\begin{aligned}
\mathcal{L}^l\!=\! \mathcal{L}^{\text{CE}}_{\Rsh} + \mathcal{L}^{\text{CE}}_{\downarrow} + \mathcal{L}^{\text{CE}}_{\uparrow} + \mathcal{L}^{\text{CE}}_{\cup}.
\end{aligned}
\vspace{-12pt}
\end{equation}

\subsection{Implementation Details}
\label{sec:network}
\vspace{-4pt}
\noindent\textbf{Backbone Network.} Our feature extraction network $F_{\text{B}}$ in \autoref{eq:imageembedding} uses the convolutional blocks of ResNet101~\cite{he2016deep}. %, which is well known for its superior performance in the image classification task and strong feature learning ability.
The stride is set to 16, \ie, the resolution of the output is 1/16 of that of the input, for high computational efficiency. In addition, the ASPP module~\cite{DeepLabv3} is applied for extracting more effective features with multi-scale context. %we apply multi-grid dilation strategy (2, 4, 8) to the last convolutional block for extracting more effective features with multi-scale context.
The ASPP-enhanced feature is compressed by a $1\!\times\!1$ convolutional layer with \textit{$ReLU$} activation. The compressed 512-\textit{d} feature is further $\times$2 upsampled and element-wisely added with the feature from the second convolutional block of ResNet101, to encode more spatial details. Thus, given an input image $I$ with a size of $K\!\times\!K$, the feature extraction network $B$ produces a new image representation $h_I\!\in\!\mathbb{R}^{\frac{K}{8}\times \frac{K}{8}\times 512}$.%, which is further used in our hierarchical neural information fusion as the image representation.

\noindent\textbf{Direct Inference Network.} We implement $F^l_{\text{LSF}}$ (\autoref{eq:levelembedding}) using a $3\!\times\!3$ convolutional layer with Batch Normalization (BN) and \textit{$ReLU$} activation, whose parameters are shared by all the nodes located in the $l$-th level. This is used for extracting specific features $\{{h}^1_{\text{LSF}}, {h}^2_{\text{LSF}}, {h}^3_{\text{LSF}}\}$ for the three semantic-levels. For each node $v$, an independent SE~\cite{senet} block, $F^v_{\text{SE}}$ in \autoref{eq:seembedding}, is further applied to extract its specific embedding $\mathbf{h}_v\!\in\!\mathbb{R}^{\frac{K}{8}\times \frac{K}{8}\times 512}$ with an extremely light-weight architecture. Then, $F_{\Rsh}$ in \autoref{eq:flsh} is implemented by a stack of three $1\!\times\!1$ convolutional layers.

\noindent\textbf{Top-down/Bottom-up Inference Network.} The architectures of the top-down $F_{\downarrow}$ (\autoref{eq:top-down}) and bottom-up $F_{\uparrow\!}$ (\autoref{eq:bottom-up2}) inference networks are very similar, and only differ in their strategies of processing the input features (see \autoref{eq:bottom-up1}). Both are achieved by three cascaded convolutional layers, with convolution sizes of $3\!\times\!3$, $3\!\times\!3$ and $1\!\times\!1$, respectively.

\noindent\textbf{Information Fusion Network.}
% $F_{\cup}$ in \autoref{eq:fusion} is implemented by a $1\!\times\!1$ convolution layer for final prediction.
$F_{\cup}$ in \autoref{eq:fusion} consists of three $1\!\times\!1$ convolutional layers with \textit{ReLU} activations for non-linear mapping.%, where the first two aim to aggregate the information from different sources, while the final one is to generate the final prediction.

\vspace{-4pt}
%------------------------------------------------------------------------
\section{Experiments}
\label{sec:exp}%
\vspace{-4pt}
%In this section, we describe our experimental setting (\autoref{sec:dataset}), present implementation details (\autoref{sec:imp}), report quantitative results comparing to several state-of-the-arts on five datasets ($\sim$20K testing images in total, \autoref{sec:qresults1}), show qualitative results (\autoref{sec:qresults2}), and study the impact of different components of our model (\autoref{sec:ablation}). More quantitative and qualitative results are provided in the supplementary material. %All the visual results shown in this section are drawn from the test sets.Additional

\subsection{Experimental Settings}
\label{sec:dataset}
\vspace{-4pt}
\noindent\textbf{Datasets:} We perform extensive experiments on the following four widely-tested datasets:
%benchmarks, namely LIP~\cite{gong2017look}, PASCAL-Person-Part~\cite{xia2017joint}, PPSS~\cite{luo2013pedestrian}, ATR~\cite{liang2015deep} and Fashion Clothing~\cite{Luo_2018_TGPnet}.
\begin{itemize}[leftmargin=*]
\setlength{\itemsep}{0pt}
\setlength{\parsep}{-2pt}
\setlength{\parskip}{-0pt}
\setlength{\leftmargin}{-10pt}
\vspace{-4pt}%In particular, semantic
\item\textbf{LIP~\cite{gong2017look}} has 50,462 single-person images with elaborate pixel-wise annotations of 19 part categories (\eg, \textit{hair}, \textit{face}, \textit{left-/right-arms}, \textit{left-/right-legs}, \textit{left-/right-shoes}, \etc). %LIP images are collected from realistic scenarios containing people appearing with challenging poses, viewpoints, \etc.
% LIP images are collected from realistic scenarios and split into 30,462 for training, 10,000 for validation and 10,000 for testing.
The images are divided into 30,462 samples for training, 10,000 for validation and 10,000 for testing.
%\item\textbf{LIP~\cite{gong2017look}} has 50,462 single-person images with  pixel-wise annotations of 19 part categories (\ie, \textit{hat}, \textit{hair}, \textit{gloves}, \textit{sunglasses}, \textit{upper-clothes}, \textit{dress}, \textit{coat}, \textit{socks}, \textit{pants}, \textit{jumpsuits}, \textit{scarf}, \textit{skirt}, \textit{face}, \textit{left-/right-arms}, \textit{left-/right-legs}, \textit{left-/right-shoes}). %LIP images are collected from realistic scenarios containing people appearing with challenging poses, viewpoints, \etc.
%% LIP images are collected from realistic scenarios and split into 30,462 for training, 10,000 for validation and 10,000 for testing.
%All images are divided into 30,462 for training, 10,000 for validation and 10,000 for testing.

\item\textbf{PASCAL-Person-Part~\cite{xia2017joint}} contains multiple humans
per image in unconstrained poses and occlusions (1,716 for training and 1,817 for testing). It provides careful pixel-wise annotations for six body parts (\ie, \textit{head,
torso, upper-/lower-arms}, and \textit{upper-/lower-legs}).

\item\textbf{ATR~\cite{liang2015deep}} includes 7,700 images (6,000 for training, 700 for validation and 1,000 for testing), annotated at pixel-level with
17 categories, \eg, \textit{hat, sunglass, face, upper-clothes, pants, left-/right-arms, left-/right-legs}, \etc.
%\ie, \textit{hat, hair, sunglass, face, upper-clothes, skirt, pants, dress, belt, left-/right-arms, left-/right-legs, left-/right-shoes, bag,} and \textit{scarf}.

\item\textbf{Fashion Clothing~\cite{Luo_2018_TGPnet}} consists of Colorful Fashion Parsing~\cite{liu2014fashion}, Fashionista~\cite{yamaguchi2012parsing}, and  Clothing Co-Parsing~\cite{yang2014clothing}. It is more concerned with human clothing details, including 17 categories (\eg, \textit{glass, hair, pants, shoes, shirt, upper-clothes, skirt, scarf, socks}, \etc). It has 4,371 images in total (3,934 for training, and 437 for testing).

    % (\ie, \textit{jewelry, bag, coat, belt, dress, glass, hair, pants, shoes, shirt, skin, skirt, upper-clothes, underwear, scarf, socks,} and \textit{hat}). It has 4,371 images in total (3,934 for training, and 437 for testing).

%\item\textbf{PPSS~\cite{luo2013pedestrian}} has 3,673 samples, collected from 171 surveillance videos containing diverse general challenges (\ie, occlusion, illumination variation) in real-word scenes.
%% PPSS is divided into a training set of 1,781 images and a testing set of 1,892 images.%
%PPSS is divided into 1,781 and 1,892 images for training and testing, respectively.
%Pixel-wise annotations for \textit{hair, face, upper-/lower-clothes, arm}, and \textit{leg} are provided.
\vspace{-4pt}
\end{itemize}

\noindent\textbf{Evaluation Metrics:} For LIP, following its standard protocol~\cite{zhao2017self}, we report pixel accuracy, mean accuracy and mean Intersection-over-Union (mIoU). %See for detailed definition of these metrics.
For PASCAL-Person-Part, following conventions~\cite{xia2016zoom,xia2017joint,luo2018macro}, the performance is evaluated in terms of mIoU. For ATR and Fashion Clothing, we report five metrics as~\cite{Luo_2018_TGPnet} does, including pixel accuracy, foreground accuracy, average precision, average recall, and average F1-score.

\begin{table}[t]
\centering\small
\begin{threeparttable}
\setlength\tabcolsep{4pt}
\renewcommand\arraystretch{1.00}
\resizebox{0.4\textwidth}{!}{
\begin{tabular}{r||c|c|c}    % {lccc}
\hline\thickhline
\rowcolor{mygray}
Methods~~~~~ &~~pixAcc.~~ &Mean Acc. &~Mean IoU~\\
\hline
\hline
SegNet~\cite{badrinarayanan2017segnet} &69.04 &24.00 &18.17\\
FCN-8s~\cite{long2015fully} &76.06 &36.75 &28.29\\
DeepLabV2~\cite{chen2018deeplab} &82.66 &51.64 &41.64\\
Attention~\cite{chen2016attention} &83.43 &54.39 &42.92\\
Attention+SSL~\cite{gong2017look} &84.36 &54.94 &44.73\\
ASN~\cite{luc2016semantic} &- &- &45.41\\
SSL~\cite{gong2017look} &- &- &46.19\\
MMAN~\cite{luo2018macro}  &- &- &46.81\\
SS-NAN~\cite{zhao2017self} &87.59 &56.03 &47.92\\
MuLA~\cite{nie2018mutual} &\textbf{88.5} &60.5 &49.3\\
CE2P~\cite{CE2P2019} &87.37 &63.20 &53.10\\ \hline
Ours &88.03 &\textbf{68.80} &\textbf{57.74} \\ \hline
\end{tabular}
}
\end{threeparttable}
\vspace{-8pt}
\caption{\small \textbf{Comparison of pixel accuracy, mean accuracy and mIoU on LIP \texttt{val}~\cite{gong2017look}}.  (Higher values are better. The best score is marked in \textbf{bold}. These notes are the same for other tables.)}
\label{tab:LIP1}
\vspace{-12pt}
\end{table}

\begin{table*}[t]
\centering\small
\begin{threeparttable}
\setlength\tabcolsep{1pt}
\renewcommand\arraystretch{1.00}
\resizebox{\textwidth}{!}{
\begin{tabular}{r||cccccccccccccccccccc|c}    % {lccc}
\hline\thickhline
\rowcolor{mygray}
Methods~~~~~~~~&Hat &Hair &Glov &Sung &Clot &Dress &Coat &Sock &Pant &Suit &Scarf &Skirt &Face &L-Arm &R-Arm &L-Leg &R-Leg &L-Sh &R-Sh &B.G. &Ave.\\
\hline
\hline
SegNet~\cite{badrinarayanan2017segnet} &26.60&44.01 &0.01 &0.00 &34.46 &0.00 &15.97 &3.59 &33.56 &0.01 &0.00 &0.00 &52.38 &15.30 &24.23 &13.82&13.17 &9.26 &6.47 &70.62 &18.17\\
FCN-8s~\cite{long2015fully} &39.79&58.96 &5.32 &3.08 &49.08&12.36&26.82&15.66&49.41 &6.48 &0.00 &2.16 &62.65 &29.78 &36.63 &28.12&26.05&17.76&17.70&78.02 &28.29\\
DeepLabV2~\cite{chen2018deeplab} &56.48&65.33&29.98&19.67&62.44&30.33&51.03&40.51&69.00&22.38&11.29&20.56&70.11 &49.25 &52.88 &42.37&35.78&33.81&32.89&84.53 &41.64\\
Attention~\cite{chen2016attention} &58.87&66.78&23.32&19.48&63.20&29.63&49.70&35.23&66.04&24.73&12.84&20.41&70.58 &50.17 &54.03 &38.35&37.70&26.20&27.09&84.00 &42.92\\
Attention+SSL~\cite{gong2017look} &59.75&67.25&28.95&21.57&65.30&29.49&51.92&38.52&68.02&24.48&14.92&24.32&71.01 &52.64 &55.79 &40.23&38.80&28.08&29.03&84.56 &44.73\\
ASN~\cite{luc2016semantic} &56.92&64.34&28.07&17.78&64.90&30.85&51.90&39.75&71.78&25.57 &7.97 &17.63&70.77 &53.53 &56.70 &49.58&48.21&34.57&33.31&84.01 &45.41\\
SSL~\cite{gong2017look} &58.21&67.17&31.20&23.65&63.66&28.31&52.35&39.58&69.40&28.61&13.70&22.52&74.84 &52.83 &55.67 &48.22&47.49&31.80&29.97&84.64 &46.19\\
MMAN~\cite{luo2018macro} &57.66&65.63&30.07&20.02&64.15&28.39&51.98&41.46&71.03&23.61 &9.65 &23.20&69.54 &55.30 &58.13 &51.90&52.17&38.58&39.05&84.75 &46.81\\
SS-NAN~\cite{zhao2017self} &63.86 &70.12&30.63&23.92&70.27&33.51&56.75&40.18&72.19&27.68&16.98&26.41&75.33&55.24&58.93&44.01&41.87&29.15&32.64
&\textbf{88.67} &47.92\\
CE2P~\cite{CE2P2019} &65.29 &72.54 &39.09 &32.73 &69.46 &32.52 &56.28 &49.67 &74.11 &27.23 &14.19 &22.51 &75.50 &65.14 &66.59 &60.10 &58.59 &46.63 &46.12 &87.67 &53.10\\ \hline
Ours &\textbf{69.55} &\textbf{73.45} &\textbf{45.17} &\textbf{41.45} &\textbf{70.57} &\textbf{38.52} &\textbf{57.94} &\textbf{54.02} &\textbf{75.07} &\textbf{28.00} &\textbf{31.92} &\textbf{30.20} &\textbf{76.38} &\textbf{68.28} &\textbf{69.49} &\textbf{65.52} &\textbf{65.51} &\textbf{52.67} &\textbf{53.38} &87.99 &\textbf{57.74}\\ \hline
\end{tabular}
}
\end{threeparttable}
\vspace{-8pt}
\caption{\small \textbf{Per-class comparison of mIoU with state-of-the-art methods  on  LIP \texttt{val}~\cite{gong2017look}}.}
\label{tab:LIP2}
\vspace{-16pt}
\end{table*}

\noindent\textbf{Training Settings:} During training, the weights of the backbone network are loaded from ResNet101~\cite{he2016deep} pre-trained on ImageNet~\cite{ImageNet}, and the remaining layers are randomly initialized.
%Different from~\cite{chen2018deeplab}, the stride of the backbone network is set to 16 (the resolution of the outputs of ResNet101 is 1/16 of inputs' resolution, 1/8 in~\cite{chen2018deeplab}), dilation strategy is only applied to the last block of ResNet101 (multi-grid dilation strategy (2, 4, 8)).
For data preparation, following~\cite{CE2P2019,gong2018instance}, we apply data augmentation techniques for all the training data, including randomly scaling, cropping and left-right flipping. The random scale is set from 0.5 to 2.0, while the crop size is set to $473\!\times\!473$.
For optimization, we adopt SGD with a momentum of 0.9, and weight\_decay of 0.0005. % with a momentum optimizer, where momentum is 0.9, and weight\_decay is 0.0005.
For the learning rate, we use the `poly' learning rate schedule~\cite{chen2018deeplab,zhao2017pspnet}, $lr\!=\!base\_lr\!\times\!(1\!-\!\frac{iters}{total\_iters})^{power}$, in which \textit{power}$=$$0.9$ and \textit{base\_lr}$=$$0.007$. The \textit{total\_iters} is %equal to
$epochs\!\times\!batch\_size$, where \textit{batch\_size}$=$$40$~and~\textit{epochs}$=$$150$. % for LIP and ATR datasets, 200 for PascalPersonPart and PPSS datasets
%For data augmentation, we adopt a random mirror and random scale between 0.5 and 2 for all datasets.
%The random scale is 0.5 to 1.5 for LIP dataset, 0.5 to 2.0 for others.
%In addition, o
We use multiple GPUs for the consumption of the large \textit{batch\_size}, and implement Synchronized Cross-GPU BN. %to optimize the parameters of BN layers.

\noindent\textbf{Testing Phase:}  %During testing, the resolution of every input is consistent with the original image.
Following general protocol~\cite{zhao2017pspnet,nie2018mutual}, we average the per-pixel classification scores at multiple scales with flipping, \ie, the scale is 0.5 to 1.5 (in increments of 0.25) times the original size. Our model does not require any other pre-/post-processing steps (\ie, over-segmentation~\cite{liang2016semantic,liang2017interpretable}, human pose~\cite{xia2017joint}, CRF~\cite{xia2017joint}), and thus achieves a processing speed of 23.0fps, averaged on PASCAL-Person-Part, which is faster than previous deep human parsers, such as Joint~\cite{xia2017joint} (0.1fps), Attention+SSL~\cite{gong2017look} (2.0fps), MMAN~\cite{luo2018macro} (3.5fps) and MuLA~\cite{nie2018mutual} (15fps).

\noindent\textbf{Reproducibility:} Our method is implemented on PyTorch and trained on four NVIDIA Tesla V100 GPUs with a 32GB memory per-card.  All the testing procedures are carried out on a single NVIDIA TITAN Xp GPU with 12GB memory for a fair speed comparison. To provide full details of our training and testing processes, we release our code in \url{https://github.com/ZzzjzzZ/CompositionalHumanParsing}.% and models

\vspace{-4pt}
\subsection{Quantitative Results}
\label{sec:qresults1}
\vspace{-4pt}

We compare the proposed method with several strong baselines on the four aforementioned challenging datasets.

\noindent\textbf{LIP~\cite{gong2017look}:} We compare our method with 11 state-of-the-arts
on LIP \texttt{val} set in \autoref{tab:LIP1}. %We can find the
Our method achieves a huge boost in average IoU (4.64\% better than the second best method, CE2P~\cite{CE2P2019} and 8.4\% better
than the third best, MuLA~\cite{nie2018mutual}). To verify its effectiveness in detail, we report per-class IoU in \autoref{tab:LIP2}. Our model improves the performance over almost all classes, especially for the ones typically associated with small regions (\eg, \textit{gloves, sunglasses, socks, shoes}), due to our top-down inference strategy. The results are also impressive for \textit{arms, legs}, and \textit{shoes}, demonstrating our model's ability to distinguish between ``left'' and ``right'' with the help of composition relations.

\begin{table}[t]
\centering\small
\begin{threeparttable}
\setlength\tabcolsep{2pt}
\renewcommand\arraystretch{1.00}
\resizebox{0.49\textwidth}{!}{
\begin{tabular}{r||ccccccc|c}    % {lccc}
\hline\thickhline
\rowcolor{mygray}
Methods~~~~~~~~~~&Head &Torso  &U-Arm &L-Arm &U-Leg &L-Leg &B.G. &Ave.\\
\hline
\hline
HAZN~\cite{xia2016zoom} &80.79 &59.11 &43.05 &42.76 &38.99 &34.46 &93.59 &56.11\\
Attention~\cite{chen2016attention} &81.47 &59.06 &44.15 &42.50 &38.28 &35.62 &93.65 &56.39\\
LG-LSTM~\cite{liang2016semantic2} &82.72 &60.99 &45.40 &47.76 &42.33 &37.96 &88.63 &57.97\\
%Joint~\cite{xia2017joint} &80.21 &61.36 &47.53 &43.94 &41.77 &38.00 &93.64 &58.06\\
Attention+SSL~\cite{gong2017look} &83.26 &62.40 &47.80 &45.58 &42.32 &39.48 &94.68 &59.36\\
Attention+MMAN~\cite{luo2018macro} &82.58 &62.83 &48.49 &47.37 &42.80 &40.40 &94.92 &59.91\\
Graph LSTM~\cite{liang2016semantic} &82.69 &62.68 &46.88 &47.71 &45.66 &40.93 &94.59 &60.16\\
SS-NAN~\cite{zhao2017self}  &86.43 &67.28 &51.09 &48.07 &44.82 &42.15 &97.23 &62.44\\
Structure LSTM~\cite{liang2017interpretable} &82.89 &67.15 &51.42 &48.72 &51.72 &45.91 &97.18 &63.57\\
Joint~\cite{xia2017joint} &85.50 &67.87 &54.72 &54.30 &48.25 &44.76 &95.32 &64.39\\
DeepLabV2~\cite{chen2018deeplab} &- &- &- &- &- &- &- &64.94\\
MuLA~\cite{nie2018mutual} &- &- &- &- &- &- &- &65.1\\
PCNet~\cite{Zhu2018ProgressiveCH} &86.81 &69.06 &55.35 &55.27 &50.21 &48.54 &96.07 &65.90\\
Holistic~\cite{li2017holistic} &- &- &- &- &- &- &- &66.3\\
WSHP~\cite{fang2018weakly} &87.15&72.28&57.07&56.21&52.43&50.36&\textbf{97.72}&67.60\\
PGN~\cite{gong2018instance} &\textbf{90.89} &\textbf{75.12} &55.83 &\textbf{64.61} &55.42 &41.57 &95.33 &68.40\\ \hline

Ours & 88.02 &72.91 &\textbf{64.31} &63.52 &\textbf{55.61} &\textbf{54.96} &96.02 &\textbf{70.76} \\ \hline
\end{tabular}
}
\end{threeparttable}
\vspace{-8pt}
\caption{\small \textbf{Per-class comparison of mIoU with state-of-the-art methods  on PASCAL-Person-Part \texttt{test}~\cite{xia2017joint}}.}
\label{tab:PASCAL-Person-Part}
\vspace{-15pt}
\end{table}

\noindent\textbf{PASCAL-Person-Part~\cite{xia2017joint}:} On its \texttt{test} set, we compare our method with 15 state-of-the-arts using IoU score. As shown in \autoref{tab:PASCAL-Person-Part}, our model outperforms previous methods across the vast majority of classes and on average.

\noindent\textbf{ATR~\cite{liang2015deep}:} \autoref{tab:ATR1} gives evaluation on ATR \texttt{test} set. %As seen,still evaluation
Our model again outperforms other competitors across most metrics. In particular, it achieves an average F-1 score of 85.51\%, which is 3.45\% better than TGPNet~\cite{Luo_2018_TGPnet} and 5.37\% better than Co-CNN~\cite{liang2015human}.

\noindent\textbf{Fashion Clothing~\cite{Luo_2018_TGPnet}:} We compare our method with five famous models on Fashion Clothing \texttt{test}, where we take
the pre-computed evaluation from~\cite{Luo_2018_TGPnet}. From \autoref{tab:Clothing}, we observe our model surpasses other competitors across all metrics by a large margin. Notably, it yields an F-1 score of 58.12\%, significantly outperforming TGPNet~\cite{Luo_2018_TGPnet} and Attention~\cite{chen2016attention} by +6.20\% and +9.44\%, respectively.

%\noindent\textbf{PPSS~\cite{luo2013pedestrian}:} Comparisons with four strong baselines (whose scores are taken
%from~\cite{luo2018macro}) on PPSS \texttt{test} set are shown~in \autoref{tab:PPSS}. Our model yields an mIoU of~60.5\%, while~the mIoUs of DL~\cite{luo2013pedestrian}, DDN~\cite{luo2013pedestrian}, ASN~\cite{luc2016semantic}, and MMAN~\cite{luo2018macro} are 35.2\%, 47.2\%, 50.7\%, and 52.1\%, respectively. This demonstrates our superior performance again.

Overall, our model consistently obtains promising results over different datasets, which clearly demonstrates its superior performance and strong generalizability. This also distinguishes our model from several previous state-of-the-art deep human parsers, such as~\cite{gong2017look,xia2017joint,fang2018weakly,nie2018mutual},  since it does not use extra pose annotations during training.

\begin{table}[t]
\centering\small
\begin{threeparttable}
\setlength\tabcolsep{2pt}
\renewcommand\arraystretch{1.00}
\resizebox{0.49\textwidth}{!}{
\begin{tabular}{r||cccc|c}    % {lccc}
\hline\thickhline
\rowcolor{mygray}
Methods  &~~~pixAcc.~~~  &F.G. Acc. &~~~Prec.~~~ &~~~Recall~~~ &~~~F-1~~~\\
\hline
\hline
Yamaguchi~\cite{yamaguchi2012parsing} &84.38 &55.59 &37.54 &51.05 &41.80 \\
Paperdoll~\cite{yamaguchi2013paper} &88.96 &62.18 &52.75 &49.43 &44.76 \\
M-CNN~\cite{liu2015matching} &89.57 &73.98 &64.56 &65.17 &62.81 \\
ATR~\cite{liang2015deep} &91.11 &71.04 &71.69 &60.25 &64.38 \\
DeepLabV2~\cite{chen2018deeplab} &94.42 &82.93 &78.48 &69.24 &73.53 \\
PSPNet~\cite{zhao2017pspnet} &95.20 &80.23 &79.66 &73.79 &75.84 \\
Attention~\cite{chen2016attention} &95.41 &85.71 &81.30 &73.55 &77.23 \\
DeepLabV3+~\cite{chen2018deeplabv3plus} &95.96 &83.04 &80.41 &78.79 &79.49 \\
Co-CNN~\cite{liang2015human} &96.02 &83.57 &\textbf{84.95} &77.66 &80.14 \\
TGPNet~\cite{Luo_2018_TGPnet} &\textbf{96.45} &\textbf{87.91} &83.36 &80.22 &81.76 \\ \hline
Ours &96.26 &\textbf{87.91} &84.62 &\textbf{86.41} &\textbf{85.51}\\
\hline
\end{tabular}
}
\end{threeparttable}
\vspace{-8pt}
\caption{\small \textbf{Comparison of accuracy, foreground accuracy, average precision, recall and F1-score on ATR \texttt{test}~\cite{liang2015deep}. }Please see the supplementary material for per-class performance.
}
\label{tab:ATR1}
\vspace{-8pt}
\end{table}

\begin{table}[t]
\centering\small
\begin{threeparttable}
\setlength\tabcolsep{2pt}
\renewcommand\arraystretch{1.00}
\resizebox{0.49\textwidth}{!}{
\begin{tabular}{r||cccc|c}    % {lccc}
\hline\thickhline
\rowcolor{mygray}
Methods  &~~pixAcc.~~  &F.G. Acc. &~~~Prec.~~~ &~~~Recall~~~ &~~~F-1~~~\\
\hline
\hline
Yamaguchi~\cite{yamaguchi2012parsing} &81.32 &32.24 &23.74 &23.68 &22.67 \\
Paperdoll~\cite{yamaguchi2013paper} &87.17 &50.59 &45.80 &34.20 &35.13 \\
DeepLabV2~\cite{chen2018deeplab} &87.68 &56.08 &35.35 &39.00 &37.09 \\
Attention~\cite{chen2016attention} &90.58 &64.47 &47.11 &50.35 &48.68 \\
TGPNet~\cite{Luo_2018_TGPnet} &91.25 &66.37 &50.71 &53.18 &51.92 \\ \hline
Ours &\textbf{92.20} &\textbf{68.59} &\textbf{56.84} &\textbf{59.47} &\textbf{58.12} \\
\hline
\end{tabular}
}
\end{threeparttable}
\vspace{-8pt}
\caption{\small \textbf{Comparison of pixel accuracy, foreground pixel accuracy, average precision, average recall and average f1-score on Fashion Clothing \texttt{test}~\cite{Luo_2018_TGPnet}}.}
\label{tab:Clothing}
\vspace{-12pt}
\end{table}

%\begin{table}[t]
%\centering\small
%\begin{threeparttable}
%\setlength\tabcolsep{2pt}
%\renewcommand\arraystretch{1.00}
%\resizebox{0.49\textwidth}{!}{
%\begin{tabular}{r||ccccccc|c}    % {lccc}
%\hline\thickhline
%\rowcolor{mygray}
%Methods~~ &Head &Face  &U-Cloth &Arms &L-Cloth &Legs &~B.G.~ &~Ave.~\\
%\hline
%\hline
%DL~\cite{luo2013pedestrian} &22.0 &29.1 &57.3 &10.6 &46.1 &12.9 &68.6 &35.2\\
%DDN~\cite{luo2013pedestrian} &35.5 &44.1 &68.4 &17.0 &61.7 &23.8 &80.0 &47.2\\
%ASN~\cite{luc2016semantic} &51.7 &51.0 &65.9 &29.5 &52.8 &20.3 &83.8 &50.7\\
%MMAN~\cite{luo2018macro} &53.1 &50.2 &69.0 &29.4 &55.9 &21.4 &85.7 &52.1\\ \hline
%Ours &\textbf{67.6} &\textbf{60.8} &\textbf{80.8} &\textbf{46.8} &\textbf{69.5} &\textbf{28.7} &\textbf{90.6} &\textbf{60.5} \\ \hline
%\end{tabular}
%}
%\end{threeparttable}
%\vspace{-8pt}
%\caption{\small \textbf{Comparison of mIoU on PPSS \texttt{test}~\cite{luo2013pedestrian}}.}
%\label{tab:PPSS}
%\vspace{-18pt}
%\end{table}

%\subsection{Runtime Analysis}
%
%For reference, Joint~\cite{xia2017joint}: 8s, Attention+SSL~\cite{gong2017look}: 0.5s, MMAN~\cite{luo2018macro}: 0.28s (3.57fps), SS-NAN~\cite{zhao2017self} 0.5s 2fps, Structure LSTM~\cite{liang2017interpretable} 1.3s LG-LSTM~\cite{liang2016semantic2}:0.3 Graph LSTM~\cite{liang2016semantic}: 1s, CE2P~\cite{CE2P2019}  0.034s 29fps
%Deeplab-v2~\cite{chen2018deeplab}: 0.63s 8fps (0.125+0.5)
%Attention~\cite{chen2016attention}: 0.35s
%MuLA~\cite{nie2018mutual}: (0.063s crop size 256,  15.87fps, 0.9s 11fps)
%Ours:  0.043s 23fps

%%%%%%%%%%%%%%%%%%% Figure 2%%%%%%%%%%%%%%%%%%%%%%
\begin{figure*}[t]
%%tr = 0.006, ts = 0.008
  \centering
      \includegraphics[width=1 \linewidth]{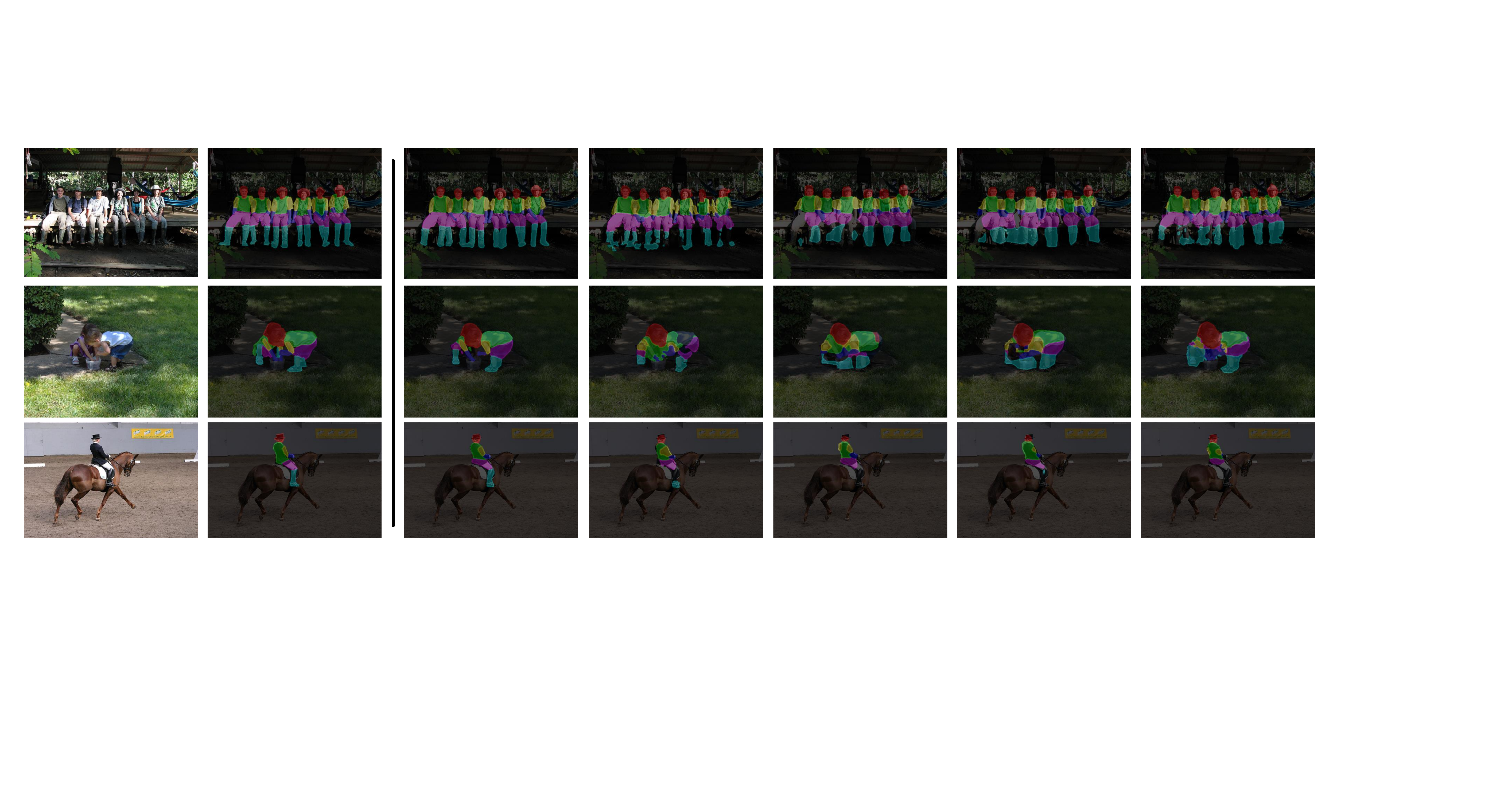}
      \vspace{0pt}
	\put(-232,10){\small(a) Image}
	\put(-176,10){\small(b) Ground-truth}
    \put(-82,10){\small(c) Ours}
    \put(-30,10){\small(d) Our backbone}
	\put(42,10){\small(e) SS-NAN~\cite{zhao2017self}}
    \put(108,10){\small(f) DeepLabV2~\cite{chen2018deeplab}}
	\put(188,10){\small(g) PGN~\cite{gong2018instance}}
\vspace{-17pt}
\caption{\small\textbf{Visual comparison results on  PASCAL-Person-Part \texttt{test} set.} Our model (c) outputs more semantically meaningful and precise predictions, compared to our backbone network (d) and other famous competitors~\cite{zhao2017self,chen2018deeplab,gong2018instance} (e-g). See \autoref{sec:qresults2} for details.}
\label{fig:visual}
\vspace{-15pt}
\end{figure*}
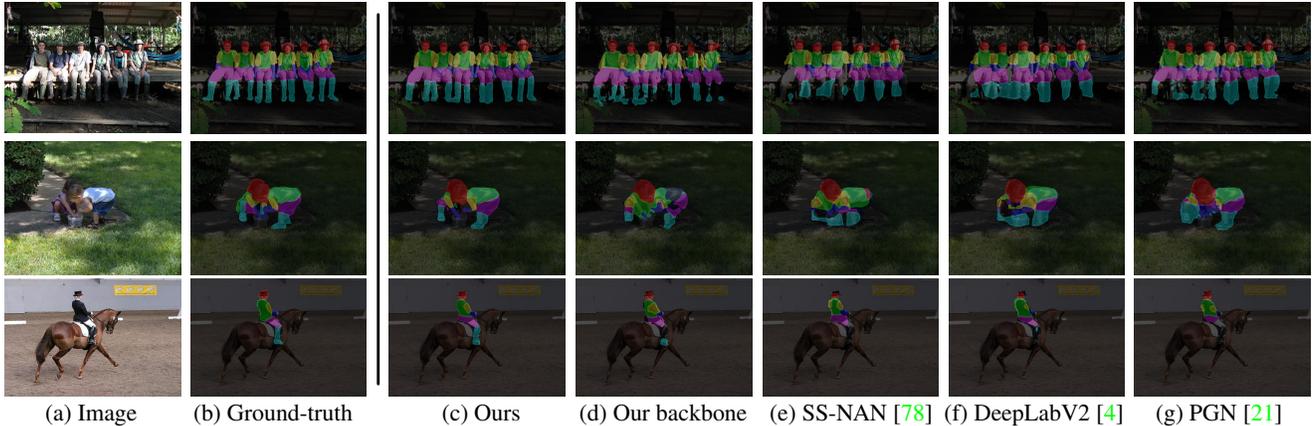

\vspace{-8pt}
\subsection{Qualitative Results}
\label{sec:qresults2}
\vspace{-4pt}
In \autoref{fig:visual}, we show some visual results on PASCAL-Person-Part \texttt{test} set. Our method yields more precise predictions compared to SS-NAN~\cite{zhao2017self}, DeepLabV2~\cite{chen2018deeplab} and PGN~\cite{gong2018instance}. For example, in the last row, our method correctly labels the lower-legs of the rider, while other methods~\cite{zhao2017self,chen2018deeplab,gong2018instance} face difficulties in this case. Our model also provides clearer details for small parts. Observed from the second row, the small lower-arm regions can be successfully segmented out with the constraint of top-down inference. In general, by effectively exploiting the human semantic hierarchy, our approach outputs reasonable results for confusing labels on the human parsing task.

%Furthermore, our model
%outputs semantically meaningful and precise predictions  despite the existence of
%large appearance and position variations.
%
%Our model estimates
%the overall part configuration more accurately. For example, in the 2 rd row of Fig. 4, we correctly labels the right
%arm of the person while the other two baseline methods label it as upper-leg and lower-leg. Furthermore, our model
%outputs semantically meaningful and precise predictions  despite the existence of
%large appearance and position variations.
%
%Our full model is able to give clearer details of arms and legs, especially for small-scale parts or the regions with similar appearances.
%For example, observed from the full-body image(e), the small regions (e.g.
%left or right shoe) can be successfully segmented out by our
%method with the constraint of top-down inference. These regions with similar appearances
%can be recognized and separated by the top-down guidance from their parent nodes.   In general, by effectively exploiting human semantic hierarchy, our approach outputs more reasonable results for confusing labels on the human parsing task.
\begin{table}[t]
\centering\small
\begin{threeparttable}
\setlength\tabcolsep{2pt}
\renewcommand\arraystretch{1.00}
\resizebox{0.48\textwidth}{!}{
\begin{tabular}{c|r||c|c|c}    % {lccc}
\hline\thickhline
\rowcolor{mygray}
 &&\multicolumn{3}{c}{mIoU} \\
  \cline{3-5}
 \rowcolor{mygray}
\multirow{-2}*{Aspects}&\multirow{-2}*{Methods~~~~~~~~~~~~~} &1st Level &2nd Level &3rd Level\\
\hline
\hline
\textbf{Full} & direct~+~bottom-up~+ &\multirow{2}*{70.76} &\multirow{2}*{81.62} &\multirow{2}*{91.31}\\
        \specialrule{0em}{-0.5pt}{-2pt}
\textbf{model} & top-down~+~conditional fusion&&\\
\hline
Backbone &direct infer. w/o hierarchy& 64.14 & - & -  \\\hline
\multirow{4}*{Variant}&direct & 65.27 & 77.83 & 88.29 \\
&{direct~+~bottom-up} & 65.42 & 78.37 & 90.10 \\
&{direct~+~top-down} & 69.02 & 78.91 & 88.40 \\
&{direct~+~bottom-up~+~top-down}&69.43 &80.34 &91.02 \\
\hline
\end{tabular}
}
\end{threeparttable}
\vspace{-8pt}
\caption{\small \textbf{Ablation study on PASCAL-Person-Part \texttt{test}~\cite{xia2017joint}}.}
\label{tab:ablation}
\vspace{-18pt}
\end{table}

\vspace{-3pt}
\subsection{Ablation Study}
\label{sec:ablation}
\begin{figure}[t]
%%tr = 0.006, ts = 0.008
  \centering
      \includegraphics[width=0.99\linewidth]{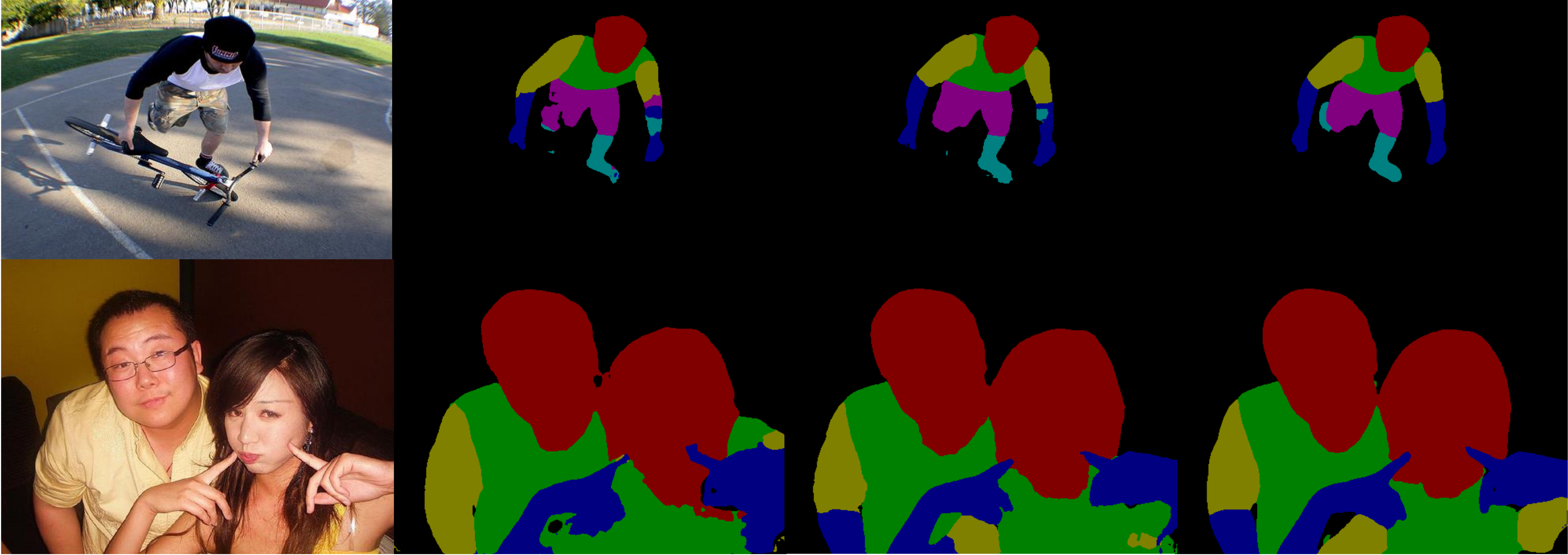}
     \mbox{}\hfill {\small (a)} \hfill\mbox{}
     \mbox{}\hfill {\small (b)} \hfill\mbox{}
     \mbox{}\hfill {\small (c)} \hfill\mbox{}
     \mbox{}\hfill {\small (d)} \hfill\mbox{}
\vspace{-8pt}
\caption{\small \textbf{Visual comparison results} from our (b) backbone, (c) compositional information fusion, and (d) full model.}
\label{fig:abs}
\vspace{-18pt}
\end{figure}

\vspace{-4pt}
%To analyze and quantify the effectiveness of each essential components of our algorithm,
\autoref{tab:ablation} shows an evaluation of our full model
compared to ablated versions without certain key components. All the variants are retrained independently with their specific network architectures. %The full model and its variants are trained with the full hierarchy, but some inference processes are disabled in the variants.
Here, 1st-Level denotes the automatic parts (\eg, \textit{head, leg}, \etc) in $\mathcal{V}^1$, 2nd-Level  $\mathcal{V}^2$ (\textit{lower-/upper body}), and 3rd-Level  $\mathcal{V}^3$ (\textit{full body}). The experiments are performed on
PASCAL-Person-Part~\cite{xia2017joint}  \texttt{test} set using mIoU metric.
Three essential conclusions can be drawn from our results. First, instead of only modeling the fine-grained parts in $\mathcal{V}^1$ (\ie, \textit{backbone}), even directly learning to parse the whole human hierarchy (\ie, \textit{direct}) can bring a performance gain (64.14$\rightarrow$65.27). This suggests that modeling the human hierarchy leads to a comprehensive understanding of human semantics.
Second, further considering bottom-up and top-down inference provides substantial performance gain, demonstrating the benefit of exploiting human structures and efficient information fusion strategies in this problem.
Note that in (\textit{direct} vs.~\textit{direct+bottom-up}) and (\textit{direct+top-down} vs.~\textit{direct+bottom-up+top-down}), even for the 1st-level nodes that do not have bottom-up inference, the training itself brings performance gain. The reason is that the bottom-up inference explicitly captures compositional relations and thus improves the quality of the learnt features. Similar observations can also be found in (\textit{direct} vs.~\textit{direct+top-down}) and  (\textit{direct+bottom-up} vs.~\textit{direct+bottom-up+top-down}) for the 3rd-level node. These observations suggest the compositional information fusion not only improves the predictions during inference but also boosts the learning ability of our human parser model.
Third, conditionally fusing information boosts performance, as the information from low-quality sources can be suppressed. This also provides a new glimpse into the information fusion mechanism over hierarchical models. A visual comparison between the results from our backbone network, our model only using compositional fusion and our full model can be found in \autoref{fig:abs} (b-d), which intuitively shows the improvements from our conditional and compositional information fusion.

\vspace{-8pt}
\section{Conclusion}
\vspace{-4pt}
In this work, we parse human parts in a hierarchical form, enabling us to capture human semantics from a more comprehensive view. We tackle this hierarchical human parsing problem through a neural information fusion framework that explores the compositional relations within human structures. It efficiently combines the information from the direct, top-down, and bottom-up inference processes while considering the reliability of each process.
Extensive quantitative and qualitative comparisons performed on five datasets demonstrate that our method outperforms the current alternatives by a large margin.

{\small\noindent\textbf{Acknowledgements} The authors thank Prof. Song-Chun Zhu and Prof. Ying Nian Wu from UCLA Statistics Department for helpful comments on this work. This work reported herein was supported in part by DARPA XAI grant N66001-17-2-4029, ARO grant W911NF-18-1-0296, CCF-Tencent Open Fund, and the National Natural Science Foundation of China (No. 61632018).}

{\small
\bibliographystyle{ieee_fullname}
\bibliography{egbib}

\begin{thebibliography}{10}\itemsep=-1pt

\bibitem{badrinarayanan2017segnet}
Vijay Badrinarayanan, Alex Kendall, and Roberto Cipolla.
\newblock Segnet: A deep convolutional encoder-decoder architecture for image
  segmentation.
\newblock {\em IEEE TPAMI}, 39(12):2481--2495, 2017.

\bibitem{battaglia2018relational}
Peter~W Battaglia, Jessica~B Hamrick, Victor Bapst, Alvaro Sanchez-Gonzalez,
  Vinicius Zambaldi, Mateusz Malinowski, Andrea Tacchetti, David Raposo, Adam
  Santoro, Ryan Faulkner, et~al.
\newblock Relational inductive biases, deep learning, and graph networks.
\newblock {\em arXiv preprint arXiv:1806.01261}, 2018.

\bibitem{chen2006composite}
Hong Chen, Zi~Jian Xu, Zi~Qiang Liu, and Song~Chun Zhu.
\newblock Composite templates for cloth modeling and sketching.
\newblock In {\em CVPR}, 2006.

\bibitem{chen2018deeplab}
Liang-Chieh Chen, George Papandreou, Iasonas Kokkinos, Kevin Murphy, and Alan~L
  Yuille.
\newblock Deeplab: Semantic image segmentation with deep convolutional nets,
  atrous convolution, and fully connected crfs.
\newblock {\em IEEE TPAMI}, 40(4):834--848, 2018.

\bibitem{DeepLabv3}
Liang-Chieh Chen, George Papandreou, Florian Schroff, and Hartwig Adam.
\newblock Rethinking atrous convolution for semantic image segmentation.
\newblock {\em CoRR}, abs/1706.05587, 2017.

\bibitem{chen2016attention}
Liang-Chieh Chen, Yi Yang, Jiang Wang, Wei Xu, and Alan~L Yuille.
\newblock Attention to scale: Scale-aware semantic image segmentation.
\newblock In {\em CVPR}, 2016.

\bibitem{chen2018deeplabv3plus}
Liang-Chieh Chen, Yukun Zhu, George Papandreou, Florian Schroff, and Hartwig
  Adam.
\newblock Encoder-decoder with atrous separable convolution for semantic image
  segmentation.
\newblock In {\em ECCV}, 2018.

\bibitem{chen2014detect}
Xianjie Chen, Roozbeh Mottaghi, Xiaobai Liu, Sanja Fidler, Raquel Urtasun, and
  Alan Yuille.
\newblock Detect what you can: Detecting and representing objects using
  holistic models and body parts.
\newblock In {\em CVPR}, 2014.

\bibitem{craik1967nature}
Kenneth James~Williams Craik.
\newblock {\em The nature of explanation}, volume 445.
\newblock CUP Archive, 1967.

\bibitem{dietterich2000ensemble}
Thomas~G Dietterich.
\newblock Ensemble methods in machine learning.
\newblock In {\em International workshop on multiple classifier systems}, 2000.

\bibitem{dong2014towards}
Jian Dong, Qiang Chen, Xiaohui Shen, Jianchao Yang, and Shuicheng Yan.
\newblock Towards unified human parsing and pose estimation.
\newblock In {\em CVPR}, 2014.

\bibitem{dong2013deformable}
Jian Dong, Qiang Chen, Wei Xia, Zhongyang Huang, and Shuicheng Yan.
\newblock A deformable mixture parsing model with parselets.
\newblock In {\em ICCV}, 2013.

\bibitem{epshtein2008image}
Boris Epshtein, Ita Lifshitz, and Shimon Ullman.
\newblock Image interpretation by a single bottom-up top-down cycle.
\newblock {\em Proceedings of the National Academy of Sciences},
  105(38):14298--14303, 2008.

\bibitem{fang2018weakly}
Hao-Shu Fang, Guansong Lu, Xiaolin Fang, Jianwen Xie, Yu-Wing Tai, and Cewu Lu.
\newblock Weakly and semi supervised human body part parsing via pose-guided
  knowledge transfer.
\newblock In {\em CVPR}, 2018.

\bibitem{fang2018learning}
Hao-Shu Fang, Yuanlu Xu, Wenguan Wang, Xiaobai Liu, and Song-Chun Zhu.
\newblock Learning pose grammar to encode human body configuration for 3d pose
  estimation.
\newblock In {\em AAAI}, 2018.

\bibitem{felzenszwalb2010object}
Pedro~F Felzenszwalb, Ross~B Girshick, David McAllester, and Deva Ramanan.
\newblock Object detection with discriminatively trained part-based models.
\newblock {\em IEEE TPAMI}, 32(9):1627--1645, 2010.

\bibitem{felzenszwalb2005pictorial}
Pedro~F Felzenszwalb and Daniel~P Huttenlocher.
\newblock Pictorial structures for object recognition.
\newblock {\em IJCV}, 61(1):55--79, 2005.

\bibitem{freund1997decision}
Yoav Freund and Robert~E Schapire.
\newblock A decision-theoretic generalization of on-line learning and an
  application to boosting.
\newblock {\em Journal of Computer and System Sciences}, pages 119--139, 1997.

\bibitem{gibson1966senses}
James~Jerome Gibson.
\newblock The senses considered as perceptual systems.
\newblock 1966.

\bibitem{gilmer2017neural}
Justin Gilmer, Samuel~S Schoenholz, Patrick~F Riley, Oriol Vinyals, and
  George~E Dahl.
\newblock Neural message passing for quantum chemistry.
\newblock In {\em ICML}, 2017.

\bibitem{gong2018instance}
Ke Gong, Xiaodan Liang, Yicheng Li, Yimin Chen, Ming Yang, and Liang Lin.
\newblock Instance-level human parsing via part grouping network.
\newblock In {\em ECCV}, 2018.

\bibitem{gong2017look}
Ke Gong, Xiaodan Liang, Dongyu Zhang, Xiaohui Shen, and Liang Lin.
\newblock Look into person: Self-supervised structure-sensitive learning and a
  new benchmark for human parsing.
\newblock In {\em CVPR}, 2017.

\bibitem{gregory1970intelligent}
Richard~Langton Gregory.
\newblock The intelligent eye.
\newblock 1970.

\bibitem{grill2014functional}
Kalanit Grill-Spector and Kevin~S Weiner.
\newblock The functional architecture of the ventral temporal cortex and its
  role in categorization.
\newblock {\em Nature Reviews Neuroscience}, 15(8):536, 2014.

\bibitem{he2016deep}
Kaiming He, Xiangyu Zhang, Shaoqing Ren, and Jian Sun.
\newblock Deep residual learning for image recognition.
\newblock In {\em CVPR}, 2016.

\bibitem{hinton1999products}
Geoffrey~E Hinton.
\newblock Products of experts.
\newblock In {\em International Conference on Artificial Neural Networks},
  1999.

\bibitem{hinton2002training}
Geoffrey~E Hinton.
\newblock Training products of experts by minimizing contrastive divergence.
\newblock {\em Neural computation}, pages 1771--1800, 2002.

\bibitem{hornik1989multilayer}
Kurt Hornik, Maxwell Stinchcombe, and Halbert White.
\newblock Multilayer feedforward networks are universal approximators.
\newblock {\em Neural networks}, pages 359--366, 1989.

\bibitem{senet}
Jie Hu, Li Shen, and Gang Sun.
\newblock Squeeze-and-excitation networks.
\newblock In {\em CVPR}, 2018.

\bibitem{jain2016structural}
Ashesh Jain, Amir~R Zamir, Silvio Savarese, and Ashutosh Saxena.
\newblock Structural-{RNN}: Deep learning on spatio-temporal graphs.
\newblock In {\em CVPR}, 2016.

\bibitem{joachims2009cutting}
Thorsten Joachims, Thomas Finley, and Chun-Nam~John Yu.
\newblock Cutting-plane training of structural svms.
\newblock {\em Machine Learning}, 77(1):27--59, 2009.

\bibitem{khaleghi2013multisensor}
Bahador Khaleghi, Alaa Khamis, Fakhreddine~O Karray, and Saiedeh~N Razavi.
\newblock Multisensor data fusion: A review of the state-of-the-art.
\newblock {\em Information fusion}, 14(1):28--44, 2013.

\bibitem{kimchi1992primacy}
Ruth Kimchi.
\newblock Primacy of wholistic processing and global/local paradigm: a critical
  review.
\newblock {\em Psychological bulletin}, page~24, 1992.

\bibitem{ladicky2013human}
Lubor Ladicky, Philip~HS Torr, and Andrew Zisserman.
\newblock Human pose estimation using a joint pixel-wise and part-wise
  formulation.
\newblock In {\em CVPR}, 2013.

\bibitem{li2019crowdpose}
Jiefeng Li, Can Wang, Hao Zhu, Yihuan Mao, Hao-Shu Fang, and Cewu Lu.
\newblock Crowdpose: Efficient crowded scenes pose estimation and a new
  benchmark.
\newblock In {\em CVPR}, 2019.

\bibitem{li2017holistic}
Qizhu Li, Anurag Arnab, and Philip~HS Torr.
\newblock Holistic, instance-level human parsing.
\newblock {\em arXiv preprint arXiv:1709.03612}, 2017.

\bibitem{LiangS17}
Shiyu Liang and R. Srikant.
\newblock Why deep neural networks for function approximation?
\newblock In {\em ICLR}, 2017.

\bibitem{liang2017interpretable}
Xiaodan Liang, Liang Lin, Xiaohui Shen, Jiashi Feng, Shuicheng Yan, and Eric~P
  Xing.
\newblock Interpretable structure-evolving lstm.
\newblock In {\em CVPR}, 2017.

\bibitem{liang2015deep}
Xiaodan Liang, Si Liu, Xiaohui Shen, Jianchao Yang, Luoqi Liu, Jian Dong, Liang
  Lin, and Shuicheng Yan.
\newblock Deep human parsing with active template regression.
\newblock {\em IEEE TPAMI}, 37(12):2402--2414, 2015.

\bibitem{liang2016semantic}
Xiaodan Liang, Xiaohui Shen, Jiashi Feng, Liang Lin, and Shuicheng Yan.
\newblock Semantic object parsing with graph lstm.
\newblock In {\em ECCV}, 2016.

\bibitem{liang2016semantic2}
Xiaodan Liang, Xiaohui Shen, Donglai Xiang, Jiashi Feng, Liang Lin, and
  Shuicheng Yan.
\newblock Semantic object parsing with local-global long short-term memory.
\newblock In {\em CVPR}, 2016.

\bibitem{liang2015human}
Xiaodan Liang, Chunyan Xu, Xiaohui Shen, Jianchao Yang, Si Liu, Jinhui Tang,
  Liang Lin, and Shuicheng Yan.
\newblock Human parsing with contextualized convolutional neural network.
\newblock In {\em ICCV}, 2015.

\bibitem{liu2014fashion}
Si Liu, Jiashi Feng, Csaba Domokos, Hui Xu, Junshi Huang, Zhenzhen Hu, and
  Shuicheng Yan.
\newblock Fashion parsing with weak color-category labels.
\newblock {\em TMM}, 16(1):253--265, 2014.

\bibitem{liu2015matching}
Si Liu, Xiaodan Liang, Luoqi Liu, Xiaohui Shen, Jianchao Yang, Changsheng Xu,
  Liang Lin, Xiaochun Cao, and Shuicheng Yan.
\newblock Matching-cnn meets knn: Quasi-parametric human parsing.
\newblock In {\em CVPR}, 2015.

\bibitem{liu2018cross}
Si Liu, Yao Sun, Defa Zhu, Guanghui Ren, Yu Chen, Jiashi Feng, and Jizhong Han.
\newblock Cross-domain human parsing via adversarial feature and label
  adaptation.
\newblock In {\em AAAI}, 2018.

\bibitem{long2015fully}
Jonathan Long, Evan Shelhamer, and Trevor Darrell.
\newblock Fully convolutional networks for semantic segmentation.
\newblock In {\em CVPR}, 2015.

\bibitem{luc2016semantic}
Pauline Luc, Camille Couprie, Soumith Chintala, and Jakob Verbeek.
\newblock Semantic segmentation using adversarial networks.
\newblock In {\em NIPS-workshop}, 2016.

\bibitem{luo2013pedestrian}
Ping Luo, Xiaogang Wang, and Xiaoou Tang.
\newblock Pedestrian parsing via deep decompositional network.
\newblock In {\em ICCV}, 2013.

\bibitem{Luo_2018_TGPnet}
Xianghui Luo, Zhuo Su, Jiaming Guo, Gengwei Zhang, and Xiangjian He.
\newblock Trusted guidance pyramid network for human parsing.
\newblock In {\em ACMMM}, 2018.

\bibitem{luo2018macro}
Yawei Luo, Zhedong Zheng, Liang Zheng, Tao Guan, Junqing Yu, and Yi Yang.
\newblock Macro-micro adversarial network for human parsing.
\newblock In {\em ECCV}, 2018.

\bibitem{marcel1983conscious}
Anthony~J Marcel.
\newblock Conscious and unconscious perception: An approach to the relations
  between phenomenal experience and perceptual processes.
\newblock {\em Cognitive psychology}, 15(2):238--300, 1983.

\bibitem{mitchell1980need}
Tom~M Mitchell.
\newblock {\em The need for biases in learning generalizations}.
\newblock Department of Computer Science, Laboratory for Computer Science
  Research, 1980.

\bibitem{navon1977forest}
David Navon.
\newblock Forest before trees: The precedence of global features in visual
  perception.
\newblock {\em Cognitive psychology}, pages 353--383, 1977.

\bibitem{nie2018mutual}
Xuecheng Nie, Jiashi Feng, and Shuicheng Yan.
\newblock Mutual learning to adapt for joint human parsing and pose estimation.
\newblock In {\em ECCV}, 2018.

\bibitem{qi2018learning}
Siyuan Qi, Wenguan Wang, Baoxiong Jia, Jianbing Shen, and Song-Chun Zhu.
\newblock Learning human-object interactions by graph parsing neural networks.
\newblock In {\em ECCV}, 2018.

\bibitem{CE2P2019}
Tao Ruan, Ting Liu, Zilong Huang, Yunchao Wei, Shikui Wei, and Yao Zhao.
\newblock Devil in the details: Towards accurate single and multiple human
  parsing.
\newblock In {\em AAAI}, 2019.

\bibitem{ImageNet}
Olga Russakovsky, Jia Deng, Hao Su, Jonathan Krause, Sanjeev Satheesh, Sean Ma,
  Zhiheng Huang, Andrej Karpathy, Aditya Khosla, Michael~S. Bernstein,
  Alexander~C. Berg, and Fei{-}Fei Li.
\newblock Imagenet large scale visual recognition challenge.
\newblock {\em IJCV}, 115(3):211--252, 2015.

\bibitem{simo2014high}
Edgar Simo-Serra, Sanja Fidler, Francesc Moreno-Noguer, and Raquel Urtasun.
\newblock A high performance crf model for clothes parsing.
\newblock In {\em ACCV}, 2014.

\bibitem{AOG2009object}
Xi Song, Tianfu Wu, Yunde Jia, and Song-Chun Zhu.
\newblock Discriminatively trained and-or tree models for object detection.
\newblock In {\em CVPR}, 2013.

\bibitem{sutton2012introduction}
Charles Sutton, Andrew McCallum, et~al.
\newblock An introduction to conditional random fields.
\newblock {\em Foundations and Trends{\textregistered} in Machine Learning},
  pages 267--373, 2012.

\bibitem{tang2018deeply}
Wei Tang, Pei Yu, and Ying Wu.
\newblock Deeply learned compositional models for human pose estimation.
\newblock In {\em ECCV}, 2018.

\bibitem{tarr1998image}
Michael~J Tarr and Heinrich~H B{\"u}lthoff.
\newblock Image-based object recognition in man, monkey and machine.
\newblock {\em Cognition}, pages 1--20, 1998.

\bibitem{veit2018convolutional}
Andreas Veit and Serge Belongie.
\newblock Convolutional networks with adaptive inference graphs.
\newblock In {\em ECCV}, 2018.

\bibitem{wainwright2008graphical}
Martin~J Wainwright, Michael~I Jordan, et~al.
\newblock Graphical models, exponential families, and variational inference.
\newblock {\em Foundations and Trends{\textregistered} in Machine Learning},
  pages 1--305, 2008.

\bibitem{wang2011blocks}
Nan Wang and Haizhou Ai.
\newblock Who blocks who: Simultaneous clothing segmentation for grouping
  images.
\newblock In {\em ICCV}, 2011.

\bibitem{wang2015joint}
Peng Wang, Xiaohui Shen, Zhe Lin, Scott Cohen, Brian Price, and Alan~L Yuille.
\newblock Joint object and part segmentation using deep learned potentials.
\newblock In {\em ICCV}, 2015.

\bibitem{wang2019iterative}
Wenguan Wang, Jianbing Shen, Ming-Ming Cheng, and Ling Shao.
\newblock An iterative and cooperative top-down and bottom-up inference network
  for salient object detection.
\newblock In {\em CVPR}, 2019.

\bibitem{wang2018attentive}
Wenguan Wang, Yuanlu Xu, Jianbing Shen, and Song-Chun Zhu.
\newblock Attentive fashion grammar network for fashion landmark detection and
  clothing category classification.
\newblock In {\em CVPR}, 2018.

\bibitem{wu2011numerical}
Tianfu Wu and Song-Chun Zhu.
\newblock A numerical study of the bottom-up and top-down inference processes
  in and-or graphs.
\newblock {\em IJCV}, 93(2):226--252, 2011.

\bibitem{xia2016zoom}
Fangting Xia, Peng Wang, Liang-Chieh Chen, and Alan~L Yuille.
\newblock Zoom better to see clearer: Human and object parsing with
  hierarchical auto-zoom net.
\newblock In {\em ECCV}, 2016.

\bibitem{xia2017joint}
Fangting Xia, Peng Wang, Xianjie Chen, and Alan~L Yuille.
\newblock Joint multi-person pose estimation and semantic part segmentation.
\newblock In {\em CVPR}, 2017.

\bibitem{xia2016pose}
Fangting Xia, Jun Zhu, Peng Wang, and Alan~L Yuille.
\newblock Pose-guided human parsing by an and/or graph using pose-context
  features.
\newblock In {\em AAAI}, 2016.

\bibitem{yamaguchi2013paper}
Kota Yamaguchi, M Hadi~Kiapour, and Tamara~L Berg.
\newblock Paper doll parsing: Retrieving similar styles to parse clothing
  items.
\newblock In {\em ICCV}, 2013.

\bibitem{yamaguchi2012parsing}
Kota Yamaguchi, M~Hadi Kiapour, Luis~E Ortiz, and Tamara~L Berg.
\newblock Parsing clothing in fashion photographs.
\newblock In {\em CVPR}, 2012.

\bibitem{yang2014clothing}
Wei Yang, Ping Luo, and Liang Lin.
\newblock Clothing co-parsing by joint image segmentation and labeling.
\newblock In {\em CVPR}, 2014.

\bibitem{zhao2017pspnet}
Hengshuang Zhao, Jianping Shi, Xiaojuan Qi, Xiaogang Wang, and Jiaya Jia.
\newblock Pyramid scene parsing network.
\newblock In {\em CVPR}, 2017.

\bibitem{zhao2018understanding}
Jian Zhao, Jianshu Li, Yu Cheng, Terence Sim, Shuicheng Yan, and Jiashi Feng.
\newblock Understanding humans in crowded scenes: Deep nested adversarial
  learning and a new benchmark for multi-human parsing.
\newblock In {\em ACMMM}, 2018.

\bibitem{zhao2017self}
Jian Zhao, Jianshu Li, Xuecheng Nie, Fang Zhao, Yunpeng Chen, Zhecan Wang,
  Jiashi Feng, and Shuicheng Yan.
\newblock Self-supervised neural aggregation networks for human parsing.
\newblock In {\em CVPR-workshop}, 2017.

\bibitem{zheng2015conditional}
Shuai Zheng, Sadeep Jayasumana, Bernardino Romera-Paredes, Vibhav Vineet,
  Zhizhong Su, Dalong Du, Chang Huang, and Philip~HS Torr.
\newblock Conditional random fields as recurrent neural networks.
\newblock In {\em ICCV}, 2015.

\bibitem{zhou2018adaptive}
Qixian Zhou, Xiaodan Liang, Ke Gong, and Liang Lin.
\newblock Adaptive temporal encoding network for video instance-level human
  parsing.
\newblock In {\em ACMMM}, 2018.

\bibitem{Zhu2018ProgressiveCH}
Bingke Zhu, Yingying Chen, Ming Tang, and Jinqiao Wang.
\newblock Progressive cognitive human parsing.
\newblock In {\em AAAI}, 2018.

\bibitem{zhu2017your}
Shizhan Zhu, Raquel Urtasun, Sanja Fidler, Dahua Lin, and Chen Change~Loy.
\newblock Be your own prada: Fashion synthesis with structural coherence.
\newblock In {\em ICCV}, 2017.

\end{thebibliography}
}

\end{document}